\RequirePackage[hyphens]{url}

\documentclass[USenglish,oneside,twocolumn]{article}

\usepackage[utf8]{inputenc}
\usepackage[big]{dgruyter_NEW}
 
\usepackage{amsmath}
\usepackage{multirow}
\usepackage{tabularx}
\usepackage{stackengine,amssymb,graphicx}
\usepackage{xspace}
\usepackage{enumitem}
\usepackage{url}
\usepackage{array}

\newcolumntype{C}[1]{>{\centering\arraybackslash}p{#1}}

\newcommand{\sysname}{\textsf{ParChoice}\xspace}

\begin{document}

  \author*[1]{Tommi Gr\"{o}ndahl}
  \affil[1]{Aalto University, Konemiehentie 2, 02150 Espoo, Finland, E-mail: tommi.grondahl@aalto.fi}
  
  \author*[2]{N. Asokan}
  \affil[2]{University of Waterloo, 200 University Ave W, Waterloo, ON N2L 3G1, Canada, E-mail: asokan@acm.org}

\title{\huge Effective writing style transfer via combinatorial paraphrasing}

\runningtitle{Effective writing style transfer via combinatorial paraphrasing}

\begin{abstract}
{Stylometry can be used to profile or deanonymize authors against their will based on writing style. Style transfer provides a defence. Current techniques typically use either encoder-decoder architectures or rule-based algorithms. Crucially, style transfer must reliably retain original semantic content to be actually deployable. We conduct a multifaceted evaluation of three state-of-the-art encoder-decoder style transfer techniques, and show that all fail at semantic retainment. In particular, they do not produce appropriate paraphrases, but only retain original content in the trivial case of exactly reproducing the text. To mitigate this problem we propose ParChoice: a technique based on the \emph{combinatorial application of multiple paraphrasing algorithms}. ParChoice strongly outperforms the encoder-decoder baselines in semantic retainment. Additionally, compared to baselines that achieve non-negligible semantic retainment, ParChoice has superior style transfer performance. We also apply ParChoice to multi-author style imitation (not considered by prior work), where we achieve up to  $75\%$ imitation success among five authors. Furthermore, when compared to two state-of-the-art rule-based style transfer techniques, ParChoice has markedly better semantic retainment. Combining ParChoice with the best performing rule-based baseline (Mutant-X [34]) also reaches the highest style transfer success on the Brennan-Greenstadt and Extended-Brennan-Greenstadt corpora, with much less impact on original meaning than when using the rule-based baseline techniques alone. Finally, we highlight a critical problem that afflicts \emph{all} current style transfer techniques: the adversary can use the same technique for thwarting style transfer via  \textit{adversarial training}. We show that adding randomness to style transfer helps to mitigate the effectiveness of adversarial training.}
\end{abstract}

  \keywords{style transfer, style imitation, stylometry, adversarial stylometry, author profiling, profiling, deanonymization, model evasion}
%
%

\maketitle

\section{Introduction}

Freedom of speech and privacy are threatened by advances in artificial intelligence, including \textit{natural language processing} (NLP). 
In particular, \textit{stylometry} can be used to identify or profile anonymous authors based on writing style \cite{Stamatatos2009, Tempesttetal2017}.
Institutions or individuals can use stylometry to deanonymize whistle-blowers and dissidents \cite{press2018, Brennan:Greenstadt2009, Brennanetal2011}.
Deanonymization can put authors in danger of harassment \cite{Almisharietal2014} or even legal repercussions \cite{Juola2013}.
Accordingly, author deanonymization or profiling constitutes an attack on privacy \cite{Brennanetal2011, Narayanetal2012, McDonaldetal2012, Grondahl:Asokan2019}.

As a defence, the author can use \textit{style transfer}.
This process can consist of several \textit{transformations}, i.e. changes applied to the input text.
Prevalent approaches are based on \textit{encoder-decoder networks} \cite{Shenetal2017, Prabhumoyeetal2018, Shettyetal2018, Zhaoetal2018, Fuetal2018, Yangetal2018, Xuetal2018, Logeswaranetal2018}, but more traditional \textit{rule-based} techniques also continue to be used \cite{Reddy:Knight2016, Karadzhovetal2017, Mahmoodetal2019}.
Importantly, style transfer is distinguished from mere style-specific generation \cite{Huetal2017, Juutietal2018} by the requirement of \textit{semantic retainment}:
the transformed text should express equivalent content to the original.

Using both automatic and manual metrics, we conduct a detailed performance evaluation of three state-of-the-art style transfer techniques based on encoder-decoder networks \cite{Shenetal2017, Prabhumoyeetal2018, Shettyetal2018} (Sections \ref{sec:experiments}--\ref{sec:results}).
The aim of these techniques is to produce a \textit{style-neutral encoding} of the original sentence's content, and then generate the same content in the target style.
However, \textit{they all fail} at producing acceptable \textit{paraphrases} (Section \ref{sec:results-semantic}).
Semantic retainment only succeeds in the trivial case of \textit{reproducing} the input.

Such results motivate a reconsideration of alternative approaches, in particular \textit{automatic paraphrasing}.
We propose a novel style transfer technique based on \textit{combinatorial paraphrase generation}, and style specific \textit{paraphrase selection}.
The technique, which we call \textit{\sysname}, 
is inspired by prior work in rule-based style transfer (Section \ref{sec:Background}) \cite{Khosmood:Levinson2008, Khosmood:Levinson2009, Khosmood:Levinson2010, Khosmood2012, Mansoorizadehetal2016, Mihaylovaetal2016, Reddy:Knight2016, Karadzhovetal2017, Mahmoodetal2019} but involves substantial additions to existing techniques. In Section \ref{sec:methods} we discuss the paraphrasing algorithms in detail.

We compare \sysname with the three encoder-decoder baselines on four author profiling datasets derived from original work presenting the baselines (Section \ref{sec:results}).
\sysname outperforms them all in semantic retainment, especially clearly in human evaluation (Section \ref{sec:results-semantic}).
In style transfer performance, \sysname surpasses those baselines that achieve non-negligible semantic retainment (Section \ref{sec:results-imitation}).
Additionally applying \sysname to five-author style imitation, we achieve up to $75\%$ imitation success (Section \ref{sec:results-imitation}).

We also compare \sysname to two rule-based techniques that have demonstrated strong performance in prior research \cite{Karadzhovetal2017, Mahmoodetal2019} (Section \ref{sec:document-based}).
Experimenting on the Brennan-Greenstadt corpus and the Extended-Brennan-Greenstadt corpus \cite{Brennanetal2011}, we demonstrate that \sysname exerts less semantic changes than either baseline. Furthermore, even though one baseline (Mutant-X \cite{Mahmoodetal2019}) achieves a higher style transfer performance than \sysname alone, applying them both in succession significantly improves the performance of both. This combined application of \sysname and Mutant-X also retains semantics better than Mutant-X alone.

Finally, affecting \textit{all} state-of-the-art style transfer techniques (including \sysname) we highlight a serious general problem.
A strong adversary who is aware of the style transfer technique can employ \textit{adversarial training}: using the style transfer technique for adding transformed examples to the training data.

We demonstrate that adversarial training thwarts all three encoder-decoder baseline techniques \cite{Shenetal2017, Prabhumoyeetal2018, Shettyetal2018} as well as \sysname, typically with only a minor negative impact on original profiling accuracy (Section \ref{sec:results-adversarial-training}). However, adversarial training fails if paraphrase selection in \sysname is \textit{random}, which indicates that the problem can be partly mitigated by conducting the transformations randomly instead of using a specific target style.
We discuss the relation between style transfer and adversarial training, and suggest directions for future research on this problem (Sections \ref{sec:results-adversarial-training}--\ref{sec:conclusion}).

We summarize our contributions below.

\begin{itemize}

\item We present \sysname: a style transfer technique based on \textit{combinatorial paraphrasing}
(Section \ref{sec:methods}).

\item By comparing \sysname with three encoder-decoder baselines \cite{Shenetal2017, Prabhumoyeetal2018, Shettyetal2018} and two rule-based baselines \cite{Karadzhovetal2017, Mahmoodetal2019}, we demonstrate that:
\begin{itemize}
\item \sysname retains semantic information better than any baseline (Sections \ref{sec:results-semantic}, \ref{sec:document-semantic}).
\item \sysname's style transfer performance exceeds those encoder-decoder baselines that achieve non-negligible semantic retainment (Section \ref{sec:results-imitation}).
\item \sysname significantly outperforms both rule-based techniques in semantic retainment, while \sysname combined with Mutant-X \cite{Mahmoodetal2019} performs the best in style transfer (Section \ref{sec:document-based}).
\end{itemize}

\item We demonstrate that the adversary can counter style transfer by \textit{adversarial training}, except if paraphrases are selected \textit{randomly} (Section \ref{sec:results-adversarial-training}). We discuss possible reasons for this finding, and propose ways in which it can be taken into account in future work on style transfer.

\item We make the code for implementing our original contributions available.\footnote{\url{https://gitlab.com/ssg-research/mlsec/parchoice/}}

\end{itemize}
\section{Background}
\label{sec:Background}

Author attribution via stylometry has traditionally focused on standard machine learning (ML) algorithms and feature engineering \cite{Zhengetal2006, Grieve2007, Abbasi:Chen2008, Juola2012, Potthastetal2016b, Tempesttetal2017}, but deep learning methods have become more prominent in recent years \cite{Bagnall2015, Ge:Sun2016, Surendranetal2017, Brocardoetal2017}.
While there is no unanimous agreement on the most effective features \cite{Grieve2007, Juola2012, Grondahl:Asokan2019}, the \textit{Writeprints} feature set has been widely applied with success \cite{Zhengetal2006, Abbasi:Chen2008, Afrozetal2012, McDonaldetal2012, Farkhundetal2013, Almisharietal2014, Overdorf:Greenstadt2016}.
Properties beyond personal identity have also been detected from writing style, including gender and age \cite{Schleretal2006, Reddy:Knight2016}. We denote the detection of any author attribute as \textit{(author) profiling}, deanonymization being a special case. We use the term \textit{(author) profiler} for ML classifiers used for profiling.

\begin{table}[t!]
  \begin{center}
    \begin{tabular}{c|p{6cm}}
    Techniques & Transformations applied \\ \hline

    \textnormal{\cite{Khosmood:Levinson2010, Mansoorizadehetal2016}} & \textnormal{synonym replacement from WordNet} \\ \hline
    \textnormal{\cite{Reddy:Knight2016, Mahmoodetal2019}} & \textnormal{word embedding neighbour replacement} \\ \hline    
    \multirow{2}{*}{\textnormal{\cite{Khosmood:Levinson2008}}} & \textnormal{word replacement from GNU Diction \cite{GNUDiction}, hand-crafted rules} \\ \hline    
    \multirow{2}{*}{\textnormal{\cite{Castro-Castroetal2017}}} & \textnormal{synonym replacement from FreeLing \cite{Padro:Stanilovsky2012}, hand-crafted rules} \\ \hline
    \multirow{2}{*}{\textnormal{\cite{Karadzhovetal2017}}} & \textnormal{synonym/hypernym/definition replacement from WordNet or PPDB, hand-crafted rules} \\

    \end{tabular}
    \caption{Prior rule-based style transfer techniques.}
    \label{tab:rule-based}
  \end{center}
\end{table}

Mitigating author profiling requires \textit{style transfer}, i.e. transforming writing style but not semantic content.
Back-and-forth machine translation provides a simple but highly limited technique \cite{Brennanetal2011, Caliskan:Greenstadt2012, Almisharietal2014, Macketal2015, Dayetal2016}, as it does not allow targeting or avoiding any particular style.
Another classical alternative is rule-based paraphrase replacement from knowledge bases \cite{Khosmood:Levinson2008, Khosmood:Levinson2009, Khosmood:Levinson2010, Khosmood2012, Reddy:Knight2016, Karadzhovetal2017, Castro-Castroetal2017, Mahmoodetal2019}, which we expand on with \sysname. Table \ref{tab:rule-based} summarizes prior rule-based style transfer techniques.

As opposed to rule-based paraphrasing, recent style transfer research has heavily concentrated on sequence-to-sequence mapping via \textit{encoder-decoder networks} \cite{Shenetal2017, Prabhumoyeetal2018, Shettyetal2018, Zhaoetal2018, Fuetal2018, Yangetal2018, Xuetal2018, Logeswaranetal2018}.
Such techniques  aim at producing a \textit{style-neutral encoding} of the original sentence, which serves as the input to a \textit{style-specific decoder}.
Their main differences concern the training algorithms used to enforce (i) the style-neutrality of the latent encoding, (ii) the style-specificity of the decoding, (iii) and semantic retainment.
In Section \ref{sec:results}, we evaluate the performance of three state-of-the-art techniques that aim at reaching (i)--(iii) by different means \cite{Shenetal2017, Prabhumoyeetal2018, Shettyetal2018}. As our results illustrate, none attain all three simultaneously.
\section{Problem statement}
\label{sec:problem-statement}

The entities involved in style transfer are the \textit{author} and the \textit{adversary}.
The author belongs to a \textit{class} $C_1$, which is a set of authors. A special case of such a class is \textit{author identity}, which is a singleton set containing only the author.
The adversary has an \textit{author profiler} $P$, which is a ML classifier used to profile texts by author class. We denote the predicted class of a text $T$ as $P(T)$. If the author has written $T$, profiling succeeds when $P(T) = C_1$ and fails otherwise.
As a defence, the author produces a \textit{transformed} text $T^*$.
She\footnote{For notational convenience, we denote the author as ``she'' and the adversary as ``he''.} succeeds in \textit{style transfer} if $P(T^*) \neq C_1$, and succeeds in \textit{imitating} another class $C_2 \neq C_1$ if $P(T^*) = C_2$.
Style transfer and imitation are thus assimilated in two-class settings.

\noindent{\textbf{Adversary models:}}
The adversary can access labeled \textit{profiler training data}, which he uses to train the author profiler $P$. The labels include the author's true class $C_1$. The adversary also accesses a text written by the author; this being originally unknown to the adversary. He profiles the text with $P$ and receives $P$'s prediction of the text's author class. In the baseline scenario the text is $T$, i.e. the original unmodified text. In style transfer scenarios it is $T^*$, i.e. $T$ transformed by the author.

We distinguish between two adversary types. The \textit{weak} adversary has no access to the author's style transfer technique, and trusts the profiling result.
The \textit{strong} adversary knows the particular style transfer technique used by the author, and can use the same technique to transform any other text he accesses.
To thwart style transfer he can use \textit{adversarial training}, i.e. re-training the author profiler with transformed training data.

\begin{table}[t!]
  \begin{center}
    \begin{tabular}{c | C{2.7cm} C{2.7cm}}
    \multirow{2}{*}{Architecture} & \multicolumn{2}{c}{Training data} \\
    & Same & Different \\ \hline
    Same & \textnormal{Query access} & \textnormal{Architecture access} \\
    Different & \textnormal{Data access} & \textnormal{Surrogate access} \\
    \end{tabular}
    \caption{Author types based on how the author's surrogate profiler relates to the adversary's author profiler.}
    \label{tab:access-assumptions}
  \end{center}
\end{table}

\noindent{\textbf{Assumptions:}}
The author can either perform random transformations, or select transformations to avoid or target a specific style.
For the latter, she needs a \textit{surrogate profiler}, which is a ML classifier trained on \textit{surrogate training data}. We distinguish between different author types by the surrogate profiler's relation to the adversary's profiler $P$.

If the surrogate profiler is the same as $P$, the author has \textit{query access}. Alternatively, the surrogate profiler can differ from $P$ in model architecture or training data, giving her \textit{data access} or \textit{architecture access}, respectively.
Finally, the weakest author only has access to a surrogate profiler that is distinct from $P$ in both architecture and training data. This \textit{surrogate access} represents the most realistic use scenario. We summarize the different access variants in Table \ref{tab:access-assumptions}.
\begin{figure*}
\includegraphics[width=510pt]{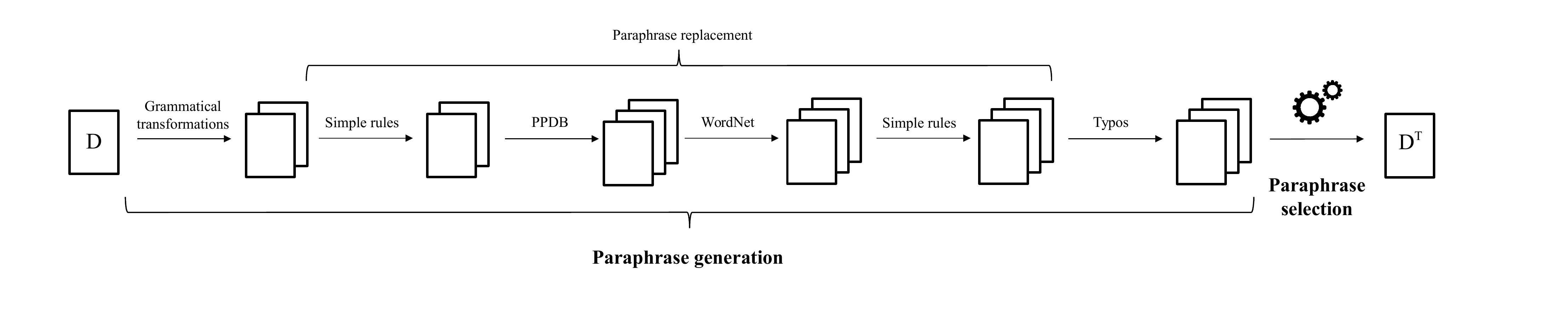}
\captionsetup{justification=centering}
\caption{ParChoice pipeline}
\label{fig:parchoice-pipeline}
\end{figure*}

\section{Design of ParChoice}
\label{sec:methods}

Figure \ref{fig:parchoice-pipeline} shows an overview of the \sysname pipeline. It consists of two stages: (i) \textit{paraphrase generation}, which takes an input document and generates a set of \textit{paraphrase candidates} (\ref{sec:paraphrase-generation}); and (ii) \textit{paraphrase selection}, which selects the candidate closest to the target writing style (\ref{sec:paraphrase-selection}).
In this section, we explain each stage.

\subsection{Paraphrase generation}
\label{sec:paraphrase-generation}

Our paraphrase generation stage consists of four modules, which we discuss in \ref{sec:grammatical-transformations}--\ref{sec:typos}.
We generate the \textit{Cartesian product} of all transformations, with the aim of producing maximally varied paraphrases.
We call this approach \textit{combinatorial paraphrasing}.

\begin{table}[b!]
  \begin{center}
    \begin{tabular}{p{3cm}|p{4.5cm}}
    
    \multicolumn{1}{p{3cm}|}{\multirow{2}{*}{\shortstack[l]{\textbf{Sentence} \\ \textbf{(category)}}}} & \multicolumn{1}{p{3cm}}{\multirow{2}{*}{\shortstack[l]{\textbf{Transformations} \\ \textbf{(category)}}}} \\
    \multicolumn{1}{p{3cm}|}{} & \multicolumn{1}{p{3cm}}{} \\ \hline
    \multicolumn{1}{p{3cm}|}{\multirow{2}{*}{\shortstack[l]{\textnormal{John saw Mary.} \\ \textnormal{(active)}}}} & \multicolumn{1}{p{3cm}}{\multirow{2}{*}{\shortstack[l]{\textnormal{Mary was seen by John.} \\ \textnormal{(passive)}}}} \\
    \multicolumn{1}{p{3cm}|}{} & \multicolumn{1}{p{3cm}}{} \\ \hline

    \multicolumn{1}{p{3cm}|}{\multirow{6}{*}{\shortstack[l]{\textnormal{John didn't see Mary.} \\ \textnormal{(negated, active)}}}} & \multicolumn{1}{p{3cm}}{\multirow{2}{*}{\shortstack[l]{\textnormal{Mary wasn't seen by John.} \\ \textnormal{(negative, passive)}}}} \\ 
    \multicolumn{1}{p{3cm}|}{} & \multicolumn{1}{p{3cm}}{} \\ \cline{2-2}
    \multicolumn{1}{p{3cm}|}{} & \multicolumn{1}{p{3cm}}{\multirow{2}{*}{\shortstack[l]{\textnormal{\textit{I don't think (that)} John saw} \\ \textnormal{Mary. (affirmed, active)}}}} \\
    \multicolumn{1}{p{3cm}|}{} & \multicolumn{1}{p{3cm}}{} \\ \cline{2-2}
    \multicolumn{1}{p{3cm}|}{} & \multicolumn{1}{p{3cm}}{\multirow{2}{*}{\shortstack[l]{\textnormal{\textit{I don't think (that)} Mary was} \\ \textnormal{seen by John. (affirmed, passive)}}}} \\
    \multicolumn{1}{p{3cm}|}{} & \multicolumn{1}{p{3cm}}{} \\ \hline
    
    \multicolumn{1}{p{3cm}|}{\multirow{6}{*}{\shortstack[l]{\textnormal{Did John see Mary?} \\ \textnormal{(question, active)}}}} & \multicolumn{1}{p{3cm}}{\multirow{2}{*}{\shortstack[l]{\textnormal{Was Mary seen by John?} \\ \textnormal{(question, passive)}}}} \\
    \multicolumn{1}{p{3cm}|}{} & \multicolumn{1}{p{3cm}}{} \\ \cline{2-2}
    \multicolumn{1}{p{3cm}|}{} & \multicolumn{1}{p{3cm}}{\multirow{2}{*}{\shortstack[l]{\textnormal{\textit{Is it (true) that} John saw Mary?} \\ \textnormal{(declarative, active)}}}} \\
    \multicolumn{1}{p{3cm}|}{} & \multicolumn{1}{p{3cm}}{} \\ \cline{2-2}
    \multicolumn{1}{p{3cm}|}{} & \multicolumn{1}{p{3cm}}{\multirow{2}{*}{\shortstack[l]{\textnormal{\textit{Is it (true) that} Mary was seen} \\ \textnormal{by John? (declarative, passive)}}}} \\
    \multicolumn{1}{p{3cm}|}{} & \multicolumn{1}{p{4.3cm}}{} \\

    \end{tabular}
    \caption{Examples of grammatical transformations \\ (added prefixes in italics).}
    
    \label{tab:grammatical-transformations}
  \end{center}
\end{table}

\subsubsection{Grammatical transformations}
\label{sec:grammatical-transformations}

Grammar is a crucial aspect of writing style, and especially important for maintaining content-neutrality in stylometry \cite{Hollingsworth2012, Hollingsworth2012b}. Yet, prior style transfer approaches have not systematically applied grammatical transformations (Section \ref{sec:Background}). A possible reason for this has been the lack of available techniques.

We used a recent tool developed by Gr\"{o}ndahl and Asokan, obtaining the same model used in the original work presenting it \cite{Eat2seq}. This technique is based on three tasks: 
(i) generating an abstract representation of the sentence (\textit{``EAT''}) derived from its \textit{dependency parse} \cite{Tesniere59};
(ii) transforming EAT according to targeted grammatical features; and
(iii) generating an English sentence from the transformed EAT via a NMT network.
The NMT network consists of an encoder and a decoder, both of which are LSTMs. It has been trained to translate EATs to English on a large corpus, consisting of $8.5$ million sentences derived from multiple open-source corpora. For further details, we refer to the original paper \cite{Eat2seq}.
All our transformations target only the \textit{main verb} of the sentence.
We explain the transformations below, and Table \ref{tab:grammatical-transformations} shows examples.

\noindent{\textbf{Voice:}}
We produced both active and passive variants of transitive verbs, which take both a subject and a direct object in the active voice. The direct object of an active clause is expressed as the subject in a passive clause, and the subject of an active clause is expressed in the passive via the preposition \textit{by}.

\noindent{\textbf{Negation:}}
In addition to using the negative particle \textit{not/n't}, an affirmative sentence can be negated by embedding it in a clause that states its falsity. We produced the affirmed version of an originally negated sentence, and wrapped it in the context: \textit{I don't think (that) (...)}.
The non-contracted variant \textit{I do not think (that) (...)} was later automatically produced in the paraphrase replacement stage (Section \ref{sec:paraphrase-replacement}).

\noindent{\textbf{Questions:}}
To paraphrase a polar (\textit{yes-no}) question, we first transformed it to a declarative variant, which we then appended to the prefix \textit{Is it (true) that (...)}. For negative questions, we additionally generated the affirmed declarative variant embedded in \textit{Is is not (true) that (...)} and \textit{Isn't it (true) that (...)}.

\subsubsection{Paraphrase replacement}
\label{sec:paraphrase-replacement}

After grammatical transformations we applied paraphrase replacement using \textit{simple rules} and two external paraphrase corpora: \textit{PPDB} \cite{Ganitkevitchetal2013} and \textit{WordNet} \cite{Wordnet95}.
We used the order \textit{simple} $\rightarrow$ \textit{PPDB} $\rightarrow$ \textit{WordNet} $\rightarrow$ \textit{simple}. The first application of simple rules increased the range of inputs for PPDB, and their re-application at the end further expanded the range of paraphrases considered for selection.
PPDB was applied before WordNet since it retained semantics better (Section \ref{sec:results-semantic}), and thus this order was less likely to propagate errors.

\noindent{\textbf{Simple rules:}}
We manually programmed a small set of rules to produce simple transformations.
First, we take the presence or absence of \textit{commas} as having only a marginal effect on interpretation, and hence allowed commas to be optionally removed.

Second, we treated the following as paraphrases:\footnote{We only produced the contraction \textit{'ve} after a pronoun in $\{$\textit{I, you, we, they}$\}$, and the contracted negation \textit{n't} after an auxiliary in $\{$\textit{is, are, was, were, have, has, had, wo} (variant of \textit{will}), \textit{must, should, need, ought, could, can, do, does, did}$\}$.}

\begin{itemize}
\item[] $\{\textit{not}, \textit{n't}\}
\{\textit{am}, \textit{'m}\} 
\{\textit{are}, \textit{'re}\}
\{\textit{have}, \textit{'ve}\}$
$\{\textit{nobody}, \textit{no-one}\}
\{\textit{anybody}, \textit{anyone}\}$
$\{\textit{somebody}, \textit{someone}\}$
\end{itemize}

Third, we replaced equivalent \textit{modal auxiliaries}.
However, some are only equivalent in either an \textit{affirmed} or a \textit{negated} context, but not both.
The context is negated if the auxiliary precedes a negation (\textit{not}/\textit{n't}), and affirmed otherwise. We therefore distinguished between equivalent modal auxiliary groups in affirmed and negated contexts. We additionally appended \textit{ought} with the preposition \textit{to} (following the negation in negated contexts), and conversely removed \textit{to} if \textit{ought} was replaced with another auxiliary.
The following sets display the equivalent modal auxiliary groups we used:

\begin{itemize}
\item[] \textbf{Affirmed context} \\
$\{$\textit{might, may, could, can}$\}$ $\{$\textit{should, ought, must}$\}$ \\ $\{$\textit{will, shall}$\}$
\item[] \textbf{Negated context} \\
$\{$\textit{can, could}$\}$ $\{$\textit{should, ought, must}$\}$ $\{$\textit{will, shall}$\}$
\end{itemize}
\hfill

\noindent{\textbf{PPDB:}}
The Paraphrase Database (PPDB) is a parallel corpus of paraphrases, annotated with additional semantic and syntactic information \cite{Ganitkevitchetal2013, Pavlicketal2015}.
PPDB-paraphrases have been used for author profiling \cite{Preotiuc-pietroetal2017}, indicating that they are relevant for writing style. However, they have been less prevalent in rule-based style transfer than WordNet (Table \ref{tab:rule-based}). This may be due to difficulties in their direct application, which we discuss below and help overcome in \sysname.

We restricted our use of PPDB to the \textit{equivalent} class, comprising $245 691$ paraphrase pairs. We derived these from the ``\textit{PPDB-TLDR}'' version.\footnote{\url{http://paraphrase.org/\#/download}.}
However, simply replacing a phrase with a random PPDB-paraphrase easily leads to ungrammaticality due to context effects.
We remedied this problem with a \textit{grammatical filter} that only allowed entries that fit the \textit{syntactic context} specified in the PPDB-entry.

Examples of PPDB-entries are shown in Table \ref{tab:PPDB}.
Single-word paraphrases include the part-of-speech (POS) tag, and multi-word paraphrases contain the syntactic context in the format [X/Y]. X describes the original phrase, and Y the phrase immediately following it in the original context. {Phrases are higher-level syntactic objects than words, and receive their grammatical status from their \textit{head} word \cite{Carnie2008}.
For example, the final row of Table \ref{tab:PPDB} is interpreted as \textit{i am sorry} being paraphrasable as \textit{i regret}, when followed by a sentence.

\begin{table}[t!]
  \begin{center}
    \begin{tabular}{ccc}
    Syntactic context & Phrase & Paraphrase \\ \hline
    \textnormal{[NN]} & \textnormal{restriction} & \textnormal{constraint} \\
    \textnormal{[VB]} & \textnormal{co-operate} & \textnormal{collaborate} \\
    \textnormal{[S/VP]} & \textnormal{i am sorry to have to} & \textnormal{i regret to} \\
    \textnormal{[S/S]} & \textnormal{i am sorry} & \textnormal{i regret} \\
    \end{tabular}
    \caption{Example PPDB entries. \\ (NN: noun, VB: verb, S: sentence, VP: verb phrase)}
    \label{tab:PPDB}
  \end{center}
\end{table}

We obtained the POS-tags and phrase structure of the original sentence with the \textit{Spacy} parser \cite{Honnibal:Johnson2015}.\footnote{\url{https://spacy.io/}}
For each word n-gram in the sentence, we detected the \textit{largest phrase immediately following it}, and used this to restrict paraphrase replacement. For single-word paraphrases we used the POS tag instead.
This \textit{grammatical filtering} algorithm drastically reduced ungrammatical paraphrases produced via PPDB-replacement, and we believe it to be useful in future work on automatic paraphrasing extending beyond style transfer.

\noindent{\textbf{WordNet:}}
As a manually built knowledge base of \textit{word senses}, WordNet \cite{Wordnet95} represents possible word meanings along with multiple semantic properties, including \textit{synonyms}. Word senses are stored as uninflected \textit{lemmas}.
While WordNet has commonly been used in rule-based style transfer \cite{Khosmood:Levinson2010, Mansoorizadehetal2016, Mihaylovaetal2016, Karadzhovetal2017, Mahmoodetal2019},
the lemma format is a major limitation in its direct application.

In contrast to prior studies, we \textit{inflected} the WordNet lemmas, significantly increasing their range of application. 
We created a dictionary from lemmas to their surface manifestations with different POS and inflection tags, deriving these from a large text corpus.\footnote{\label{fn:WordNet} We used the POS tagger of NLTK \cite{NLTK09}. We obtained the inflected lemmas from the same corpus of $8.5$ million sentences that was used for training the NMT network used for grammatical transformations \cite{Eat2seq} (Section \ref{sec:grammatical-transformations}). As the inflected variant of a lemma, we chose the most common surface form associated with a lemma-tag combination in the tagged corpus.} We then inflected synonyms of the original word based on its POS and inflection tag, and produced paraphrase candidates with each inflected synonym.
For \textit{word sense disambiguation} \cite{Navigli2009}, we used the \textit{simple Lesk} algorithm from Python's WSD library.\footnote{\url{https://github.com/alvations/pywsd}}

\subsubsection{Typos}
\label{sec:typos}

Typographical errors have demonstrated success at evading text classification \cite{Grondahletal2018, Lietal2019}.
However, given the vast number of possible misspellings and their varying effects on readability, introducing them randomly is not justifiable. We used a simple typo algorithm of optionally removing an apostrophe, as in \textit{you're}$\rightarrow$\textit{youre}. Additionally, we introduced typos that appear in the \textit{target corpus} of the surrogate training data. For obtaining these we used the Python port of SymSpell,\footnote{\url{https://github.com/wolfgarbe/SymSpell}} applying it to the target corpus and storing a dictionary from correct spellings to possible misspellings.
We additionally spell-checked the original sentence and included original typos to this typo dictionary. This allowed either retaining original typos or correcting them, depending on their effects on paraphrase selection.

Unlike other paraphrasing mechanisms, typos are \textit{reversible} via spell-checking.
The full paraphrase generation pipeline is not reversible: since a paraphrase could correspond to a very large number of possible inputs, it is practically impossible to find the correct input from the paraphrase alone.
However, as typos are an important aspect of stylistic variation \cite{Abbasi:Chen2008, Brennanetal2011, Ho:Ng2016}, removing them at pre-processing can potentially reduce the accuracy of author profiling. Our empirical results (Section \ref{sec:results-imitation}) indicate that their effectiveness at style transfer varies across datasets, but is usually not the most important factor.

\subsection{Paraphrase selection}
\label{sec:paraphrase-selection}

We selected the paraphrase candidate by a \textit{surrogate profiler}, which is a local author profiler trained on \textit{surrogate training data} (Section \ref{sec:problem-statement}, Table \ref{tab:access-assumptions}).

In sentence-level experiments (Section \ref{sec:results-sentence-based}), we chose the candidate that was assigned the \textit{highest difference between target and source class probabilities} by the surrogate profiler.
While this metric assimilates to the lowest source-class probability (or the highest target-class probability) in two-class settings, in multi-class settings these probabilities come apart. This allows us to perform \textit{imitation} instead of mere style transfer (Sections \ref{sec:problem-statement}, \ref{sec:results-imitation}).

In document-level experiments (Section \ref{sec:document-based}) we replicated a genetic algorithm -based paraphrase selection method from our best-performing rule-based baseline technique: Mutant-X \cite{Mahmoodetal2019}. This selection performs only style transfer instead of imitation. We first produced a set of candidates by paraphrasing a random sentence in the document. Following Mutant-X, we then ranked the candidates based on the probability of misclassification (using query access to the author profiler), and METEOR score with the original document. The best-performing candidates were then used as inputs for further iterations of the same process. Sections \ref{sec:rule-based-baselines} and \ref{sec:implementation-parchoice} further discuss the parameters used for Mutant-X and the document-based \sysname variant.
\section{Experiments}
\label{sec:experiments}

In this section we describe the datasets (\ref{sec:datasets}), evaluation setup (\ref{sec:evaluation}), and style transfer techniques (\ref{sec:style-transfer-techniques}) used in our experiments.
We compared \sysname with \textit{three encoder-decoder baseline techniques} on \textit{four two-class sentence-level datasets}.
We additionally applied \sysname to \textit{multi-author imitation} on a \textit{five-class sentence-level dataset}.
Finally, we compared \sysname with \textit{two rule-based baseline techniques} on \textit{four multiclass document-level datasets}.

\subsection{Datasets}
\label{sec:datasets}

The encoder-decoder baseline techniques \cite{Shenetal2017, Prabhumoyeetal2018, Shettyetal2018} use \textit{sentences} as inputs, and are only applicable to \textit{two-class} data.
The rule-based baselines \cite{Karadzhovetal2017, Mahmoodetal2019} are tailored for \textit{multi-author datasets} of \textit{multi-sentence documents}.
We used datasets specifically tailored for the baselines.
\sysname is applicable to all of them, which illustrates its flexibility across different use cases.

\subsubsection{Sentence-based datasets}
\label{sec:datasets-sentence}

We used four two-class corpora, one labeled by \textit{gender}, one by \textit{age}, and two by \textit{author identity}.
In the gender and age experiments, we replicated the setups of two encoder-decoder baselines \cite{Prabhumoyeetal2018, Shettyetal2018}.
We divided each dataset to a larger \textit{profiler training set}, a smaller \textit{surrogate training set} ($15\%$ of profiler training set size), and a \textit{test set}.
The original datasets are available on GitHub.\footnote{YG: \url{https://github.com/shrimai/Style-Transfer-Through-Back-Translation} \\ BA/AB/TO: \url{https://github.com/rakshithShetty/A4NT-author-masking/blob/master/README.md}}
Size-related information is presented in Table \ref{tab:datasets}. 

\noindent{\textbf{Yelp Gender (YG):}}
This dataset consists of restaurant reviews labeled by \textit{gender} \cite{Reddy:Knight2016}. 
It contains two training sets, which results in partially divergent data/architecture access between the baselines.
We discuss this in more detail in Section \ref{sec:encoder-decoder-baselines}.	

\noindent{\textbf{Blog Age (BA):}}
The blog dataset \cite{Schleretal2006} contains blog posts labeled by authorship, gender, and age.
We experimented on age profiling.\footnote{While we also experimented with gender, classification was heavily biased toward the \textit{female} class on original data, not achieving $>40\%$ accuracy on the \textit{male} class with any of our three classifiers.
Since the same classifiers worked much better with other datasets, this problem seemed to be due to the data itself rather than classifier architectures.
These results are in line with those of Shetty et al. \cite{Shettyetal2018} who also received much lower classification accuracy on gender than on age in the blog dataset.
We therefore focused on age in our experiments.}

\begin{table}[t!]
  \begin{center}
    \begin{tabular}{c|c|cc|c}
    
    \multirow{2}{*}{Dataset} & \multirow{2}{*}{Classes} & \multicolumn{2}{c|}{Training set size} & Test set \\
    && Profiler & Surrogate & size \\ \hline
    
    \textnormal{YG} & \textnormal{female, male} & $\textnormal{2577862}$ & $\textnormal{386678}$ & $\textnormal{10000}$ \\
    \textnormal{BA} & \textnormal{adult, teen} & $\textnormal{2637850}$ & $\textnormal{395676}$ & $\textnormal{10000}$ \\
    \textnormal{AB} & \textnormal{Alice, Bob} & $\textnormal{25319}$ & $\textnormal{3797}$ & $\textnormal{1176}$ \\    
    \textnormal{TO} & \textnormal{Trump, Obama} & $\textnormal{20860}$ & $\textnormal{3128}$ & $\textnormal{4668}$ \\

    \end{tabular}
    \caption{Sentence-based datasets (size: number of sentences).}
    \label{tab:datasets}
  \end{center}
\end{table}

\noindent{\textbf{Alice-Bob (AB):}}
We extracted two authors from BA:
a female in the age range $13-19$, and a male in the age range $23-40$. We call them \textit{Alice} and \textit{Bob}, respectively.

\noindent{\textbf{Trump-Obama (TO):}}
This dataset includes speeches by Barack Obama and Donald Trump \cite{Shettyetal2018}. We were able to improve profiling accuracy markedly by truncating the larger class (Obama) to the same size as the smaller. We used this balanced variant in our experiments.

\noindent{\textbf{Multi-class:}} We created a five-class dataset by appending AB with three additional authors from BA, introducing $7899 + 7270 + 6766$ additional training sentences.

\subsubsection{Document-based datasets}
\label{sec:datasets-document}

For document-level experiments, we used the Brennan-Greenstadt corpus (BG) and the Extended-Brennan-Greenstadt corpus (EBG) \cite{Brennanetal2011}.
These corpora have been manufactured specifically for the purposes of stylometry, and contain multiple documents by different authors.
BG contains $12$ authors and EBG $45$ authors.
We used the full BG, and replicated the test settings of Mahmood et al. \cite{Mahmoodetal2019} by using subsets of $5$ and $10$ authors from EBG. We additionally experimented on the whole EBG, giving us four datasets altogether: \textit{BG}, \textit{EBG$_5$}, \textit{EBG$_{10}$}, and \textit{EBG$_\text{45}$}.
Following Mahmood et al. \cite{Mahmoodetal2019}, we used $12$ documents from each author for training the profiler, and the rest for testing.

\subsection{Evaluation setup}
\label{sec:evaluation}

We measured the effectiveness of each technique on two fronts: \textit{semantic retainment} and \textit{style transfer}.

\subsubsection{Semantic retainment evaluation}
\label{sec:eval_semantic_retainment}

We measured semantic retainment using both automatic and manual metrics, and conducted a user study with independent evaluators for comparing \sysname to the encoder-decoder baselines.

\noindent{\textbf{Automatic evaluation:}}
We calculated the \textit{METEOR} score \cite{Banerjee:Lavie2005} between the original and transformed test sets.
METEOR is based on n-gram overlap, and additionally considers synonyms and paraphrases.
We used the METEOR implementation of the \textit{nlg-eval}\footnote{\url{https://github.com/Maluuba/nlg-eval}} package.

\noindent{\textbf{Manual evaluation:}}
We manually examined a subset of test set transformations, assessing whether they constituted acceptable paraphrases or had errors.
For sentence-based data, we evaluated $100$ random sentences from each two-class dataset ($50$ from each direction).
For document-based data, we manually evaluated those sentences that were transformed by all the techniques compared ($81$ altogether), and combined this evaluation with the rate of transformed sentences per document for each technique. This yields an estimation of how likely the techniques are to retain the original meaning of a sentence in a document, by not transforming it or by generating an appropriate paraphrase.

\noindent{\textbf{User study:}}
We conducted a user study on $20$ participants to evaluate transformations by \sysname and the encoder-decoder baselines.
Of the participants, $25\%$ were native English speakers. $55\%$ were female and $45\%$ male. $60\%$ were $20-30$ years old,  and $40\%$ were $30$ or older. $90\%$ had a university degree, most often on the Master's level ($65\%$).
In an online questionnaire, each participant was allocated a set of $20$ sentences drawn from the YG dataset. They were shown the original sentence along with transformed versions by all four imitation techniques (in a random order).
All users were given \textit{different} sentences to increase variation, resulting in $20 \times 20 \times 4 = 1600$ evaluations altogether.
To ensure the relevance of the evaluation, we only used imitations that were \textit{non-identical} with the original sentence, i.e. not exact reproductions.
Participants compared each variant to the original sentence, and rated it on a $0-5$ scale based on similarity of meaning.

\subsubsection{Style transfer evaluation}
\label{sec:eval_imitation_success}

For sentence-level datasets, we trained three author profilers that represent the state-of-the-art in stylometry (Section \ref{sec:Background}).
In document-level tests we used the best-performing profiler setup from Mahmood et al. \cite{Mahmoodetal2019}.

\noindent{\textbf{Profilers:}}
We adopted the most commonly used deep learning text classification techniques: \textit{long short-term memory networks} (LSTM) \cite{Hochreiter:Schmidhuber97} and \textit{convolutional neural networks} (CNN) \cite{LeCunetal1998}.
We used implementations from Shetty et al. \cite{Shettyetal2018} (LSTM) and Prabhumoye et al. \cite{Prabhumoyeetal2018} (CNN), which also form parts of our baseline style transfer techniques (Section \ref{sec:encoder-decoder-baselines}).
Both use words as input features.
The source codes are available on GitHub.\footnote{\url{https://github.com/shrimai/Style-Transfer-Through-Back-Translation} \\ \url{https://github.com/rakshithShetty/A4NT-author-masking/blob/master/README.md}}

\textit{Writeprints features} have exhibited strong performance in stylometry (Section \ref{sec:Background}) \cite{Zhengetal2006, Abbasi:Chen2008, Afrozetal2012, McDonaldetal2012, Farkhundetal2013, Almisharietal2014, Overdorf:Greenstadt2016}.
For our third sentence-based author profiler, we collected \textit{static} Writeprints features \cite{Zhengetal2006} and trained a \textit{multilayer perceptron} (MLP) profiler that we call \textit{WP}.
Static features apply to any user and are thus more general than \textit{dynamic} features, which include user-specific information. We used the following features: number of words, average word length, number of short words ($\leq3$), number of characters, digit percentage, uppercase character percentage, spacial character percentage, number of letters, number of digits, common character bigram/trigram percentages, number of hapax- and dislegomena, number of function words, number of POS tags, and number of punctuation markers.
Based on a comparative evaluation between five ML architectures (MLP, logistic regression, naive Bayes, decision trees, and support vector machines), MLP fared the best on our datasets with the static Writeprints features.
WP has a single hidden layer of $100$ nodes.

For document-based datasets, we replicated the test settings of Mahmood et al. \cite{Mahmoodetal2019}, who used a \textit{random forest classifier} trained on static Writeprints-features. This profiler had the highest performance in their comparative evaluation with other architectures.

\noindent{\textbf{Proofreading:}}
From $100$ sentences of YG, we manually produced additional \textit{proofread} transformations. The purpose of the proofreading was to ensure semantic retainment while changing as little of the transformation as possible. All corrections were made to the direction of the original sentence; i.e. we did not produce any novel paraphrases. This test evaluates how well the style transfer techniques are able to perform if the author secures semantic retainment by correcting the output.
It thus resembles semi-automatic style transfer frameworks like Anonymouth \cite{McDonaldetal2012} or AuthorWebs \cite{Dayetal2016}.

\noindent{\textbf{Adversarial training}}
We tested the effectiveness of thwarting style transfer via adversarial training on the AB and five-class datasets, with the LSTM profiler.
With each style transfer technique, we appended transformed variants to the original training set, and retrained the LSTM with this adversarial training set. We then measured profiling accuracy on both the original (non-transformed) test set and the transformations performed by the technique used for adversarial training.

\subsection{Style transfer techniques}
\label{sec:style-transfer-techniques}

We review the technical details of the baseline techniques (\ref{sec:encoder-decoder-baselines}--\ref{sec:rule-based-baselines}) and \sysname-variants (\ref{sec:implementation-parchoice}).

\subsubsection{Encoder-decoder Baselines}
\label{sec:encoder-decoder-baselines}

The main idea behind the encoder-decoder baseline techniques is to generate a style-neutral encoding of the original sentence, which functions as the input to a style-specific decoder. Some models are available as pre-trained, and the rest we trained ourselves. Pre-trained models and the code for training the baselines are available on the respective projects' GitHub pages (linked below).
The baselines partly differ in their access assumptions (cf. Table \ref{tab:access-assumptions}), as summarized in Table \ref{tab:assumptions}.

\noindent{\textbf{Cross-aligned autoencoder (CAE):}}
The CAE technique is a style-specific autoencoder that uses a method called \textit{cross-alignment} for calibrating encoding distributions \cite{Shenetal2017}. The \textit{encoder} produces a latent content variable from the input sentence, and the \textit{decoder} generates the target sentence from this content variable together with a target style feature.
CAE does not have query access to any of our author profilers, but has data access to every dataset except YG (explanation below).
We trained CAE for every dataset, using the project's code from GitHub.\footnote{\url{https://github.com/shentianxiao/language-style-transfer}}

\begin{table}[t!]
  \begin{center}
    \begin{tabular}{c|c|c}

    Name & Architecture access & Data access \\ \hline
    CAE & \normalfont{-} & \normalfont{BA/AB/TO} \\
    BT & \normalfont{CNN} & \normalfont{YG/BA/AB/TO} \\
    A$^4$NT & \normalfont{LSTM} & \normalfont{BA/AB/TO} \\
    \sysname-CNN & \normalfont{CNN} & \normalfont{YG/BA/AB/TO} \\
    \sysname-LSTM & \normalfont{LSTM} & \normalfont{BA/AB/TO} \\
    \sysname-WP & \normalfont{WP} & \normalfont{YG/BA/AB/TO} \\
    \sysname-LR$_\text{d}$ & \normalfont{-} & \normalfont{YG/BA/AB/TO} \\
    \sysname-LR$_\text{s}$ & \normalfont{-} & \normalfont{-} \\
    \sysname-random & \normalfont{-} & \normalfont{-} \\

    \end{tabular}
    \caption{Architecture/data access of sentence-based techniques.}
    \label{tab:assumptions}
  \end{center}
\end{table}

\noindent{\textbf{Back-translation (BT):}}
As an alternative to cross-alignment, BT produces the latent content variable with a pre-trained MT system \cite{Prabhumoyeetal2018}. The original English sentence is first translated to French, which is then encoded with the French-English encoder. An English decoder then produces the target sentence from the encoding, and separate decoders are trained to target specific styles.
Style-specificity is enforced by a CNN, which we also use as one of our author profilers (Section \ref{sec:eval_imitation_success}).

BT trained on YG is provided on the project's GitHub page.\footnote{\url{https://github.com/shrimai/Style-Transfer-Through-Back-Translation}} It has been trained with two separate datasets: one for the \textit{classifier} and another for the \textit{decoder} (Section \ref{sec:datasets}).
We used the decoder training set to train the other baselines, and the classifier training set for training the profilers.
Therefore BT has data access to YG, but CAE and A$^4$NT do not.
With other datasets we trained BT using the same training data for both the classifier and the generator.

\noindent{\textbf{A$^4$NT:}}
\textit{Adversarial Author Attribute Anonymity Neural Translation} (A$^4$NT) \cite{Shettyetal2018} is a style transfer technique based on \textit{generative adversarial networks} (GANs).
A GAN consists of two neural networks, where one (the \textit{classifier}) is trained to classify outputs generated by the other (the \textit{generator}), which in turn is trained to deceive the classifier \cite{Goodfellowetal2014}. The A$^4$NT-generator is trained to produce sentences classified as the target style by an LSTM, which we also use as one of our author profilers (Section \ref{sec:eval_imitation_success}). During training, semantic retainment is regulated by the \textit{reconstruction probability} of the original sentence via a reverse transformation.

We used pre-trained A$^4$NT-models for BA and TO,\footnote{\url{https://github.com/rakshithShetty/A4NT-author-masking/blob/master/README.md}} and trained A$^4$NT ourselves for YG and AB.
While we always used the full dataset for training the initial classifier and generator, hardware limitations required us to truncate YG during the GAN-training phrase. We used a subset of $100000$ sentences for this.

\subsubsection{Rule-based baselines}
\label{sec:rule-based-baselines}

We used two rule-based baselines for the document-level tests. The first \cite{Karadzhovetal2017} exhibited leading performance on the PAN2016 Author Obfuscation task \cite{Potthastetal2016}, and the second \cite{Mahmoodetal2019} achieved state-of-the-art results on the EBG corpus (Section \ref{sec:datasets-document}). Following Mahmood et al \cite{Mahmoodetal2019}, we call these \textit{PAN2016} and \textit{Mutant-X}.
We implemented both with code from the original projects (links below).

\noindent{\textbf{PAN2016:}}\footnote{\url{https://bitbucket.org/pan2016authorobfuscation/authorobfuscation/src/master/}} This technique \cite{Karadzhovetal2017} uses multiple hand-crafted rules along with word replacement from WordNet and PPDB. Unlike \sysname, PAN2016 does not conduct either inflection or grammatical filtering to increase the readability of the output. In addition to synonyms, WordNet-replacement also uses \textit{hypernyms} and \textit{definitions}. Additional hand-crafted rules include e.g. replacing or injecting stopwords, replacing or removing punctuation, and expanding contracted forms. PAN2016 contains its own stylometric feature set (similar to Writeprints), and calculates the \textit{average} of these features from the training corpus. It then alters the test document to shift its features closer to this average. Hence, it relies on \textit{data access} to the original training set, but does not require query access to the profiler.

\noindent{\textbf{Mutant-X:}}\footnote{\url{https://github.com/asad1996172/Mutant-X/}} This technique \cite{Mahmoodetal2019} replaces original words with their \textit{word embedding neighbours} obtained from a pre-trained Word2Vec \cite{Mikolovelat2013} model. The neighbour order is further modified to shift words of opposite sentiment (e.g. \textit{good} and \textit{bad}) away from each other \cite{Yuetal2018}. Mutant-X repeats random word replacement multiple times, applying a \textit{genetic algorithm} \cite{Goldberg1989} to keep the best performing variants after each iteration. Performance is measured as the weighted combined effect of METEOR score and how much original class probability is taken down. For calculating the latter, Mutant-X uses \textit{query access} to the author profiler.
We used the same hyperparameters as Mahmood et al. \cite{Mahmoodetal2019}: maximally $5\%$ of document words changed per run; $100$ runs per iteration; maximally $25$ iterations; and $0.25/0.75$ weights for METEOR and class probability, respectively.

\subsubsection{\sysname}
\label{sec:implementation-parchoice}

We implemented \sysname in Python 3.

\noindent{\textbf{Sentence-based variants:}}
To provide a maximally close comparison to the encoder-decoder baselines, we replicated the data/architecture access of BT and A$^4$NT by using the CNN and LSTM profilers for paraphrase selection, respectively (\sysname-CNN and \sysname-LSTM). We also experimented with black-box access to the WP profiler (\sysname-WP).
As a separate \textit{surrogate profiler}, we used a \textit{logistic regression} classifier with word unigrams as input features.
We trained two versions of the surrogate profiler: one with \textit{data access} to the targeted profiler's training data (\sysname-LR$_\text{d}$), and another trained on separate \textit{surrogate training data} (\sysname-LR$_\text{s}$).
For consistency across experiments, we always used $15\%$ of the author profiler training data size as the surrogate training data size (Section \ref{sec:datasets}; Table \ref{tab:datasets}).\footnote{In \sysname-LR$_\text{s}$ we also used the surrogate training data for obtaining typos (Section \ref{sec:typos}).
To maintain consistency in surrogate data sizes across all experiments, we used a different YG surrogate dataset for \sysname-LR$_\text{s}$ than the decoder training set used by A$^4$NT and CAE, even though both are distinct from the classifier training set (Section \ref{sec:encoder-decoder-baselines}, Table \ref{tab:assumptions}). The \sysname-LSTM variant was trained on the decoder training set, as it replicates the access properties of A$^4$NT (Table \ref{tab:assumptions}).}
Finally, we also experimented on \textit{random paraphrase selection} without any surrogate profiler (\sysname-random).
Rows $4-9$ of Table \ref{tab:assumptions} summarize sentence-based \sysname variants and their data/architecture access.

\noindent{\textbf{Genetic algorithm:}}
On document-level data, we used the same genetic algorithm as Mutant-X (Section \ref{sec:rule-based-baselines}), except that instead of changing random words in the document, each run randomly paraphrased \textit{one sentence} in the document. With the hyperparameters used, this meant that for $25$ iterations, $100$ new candidates of the document were produced by paraphrasing a single random sentence. The best candidates were then selected for further iterations, as in Mutant-X.
We replicated Mutant-X's paraphrase selection using a combination of METEOR and \textit{query access} to the targeted author profiler, with $0.25/0.75$ weights, respectively (Section \ref{sec:rule-based-baselines}).

\noindent{\textbf{\sysname + Mutant-X:}}
We additionally combined the two best-performing rule-based techniques: \sysname and Mutant-X.
We first ran \sysname, and then applied Mutant-X only to those documents that had not yet succeeded in style transfer.
This combination thus maximized the use of \sysname, and applied Mutant-X when \sysname alone was insufficient.

\noindent{\textbf{Hyperparameters:}}
Initial manual evaluation indicated that most semantic problems occurred in long sentences with multiple transformations. This motivated an upper limit to transformations per sentence. A limit based on sentence length is problematic for short sentences, where even minor transformations are percentually large. Instead, we used a constant edit distance limit of $10$, which allows large changes in short sentences but limits them in long sentences.
For computational efficiency, we also limited the number of PPDB- and WordNet replacements to $1000$ per sentence. Comparison with larger values indicated that further increasing this number had little to no effect on performance.
\section{Evaluation results}
\label{sec:results}

We present the results on experiments on the sentence-level (\ref{sec:results-sentence-based}) and document-level (\ref{sec:document-based}).
Raw data and example transformations are provided in Appendix \ref{appendix:raw_results}.

\subsection{Sentence-level experiments}
\label{sec:results-sentence-based}

ParChoice exhibits a \textit{higher semantic retainment} than any encoder-decoder baseline (\ref{sec:results-semantic}).
Its style transfer performance is higher than that of A$^4$NT, which is the only encoder-decoder baseline that achieves non-negligible semantic retainment (\ref{sec:results-imitation}).

\begin{table}[t!]
  \begin{center}
    \begin{tabular}{l|cccc}
     Technique & YG & BA & AB & TO \\ \hline
     CAE & $\textnormal{19.63}$ & $\textnormal{21.49}$ & $\textnormal{6.81}$ & $\textnormal{4.40}$ \\
     BT & $\textnormal{20.88}$ & $\textnormal{17.91}$ & $\textnormal{5.64}$ &  $\textnormal{4.62}$ \\
     A$^4$NT & $\textnormal{44.95}$ &  $\textnormal{48.98}$ & $\textnormal{22.98}$ & $\textnormal{19.81}$ \\
     \sysname-CNN & $\fbox{\textnormal{46.09}}$ & $\fbox{\textnormal{50.70}}$ & $\fbox{\textnormal{48.61}}$ & $\fbox{\textnormal{51.20}}$ \\
     \sysname-LSTM & $\textnormal{45.33}$ & $\textnormal{48.94}$ & $\textnormal{41.44}$ & $\textnormal{48.86}$ \\
     \sysname-WP & $\textnormal{45.20}$ & $\textnormal{48.72}$ & $\textnormal{41.06}$ & $\textnormal{49.84}$ \\
     \sysname-LR$_\text{d}$ & $\textnormal{46.00}$ & $\textnormal{48.59}$ & $\textnormal{43.02}$ & $\textnormal{49.17}$ \\
     \sysname-LR$_\text{s}$ & $\textnormal{45.89}$ & $\textnormal{48.83}$ & $\textnormal{45.82}$ & $\textnormal{50.27}$ \\
    \end{tabular}
    \caption{METEOR scores between original and transformed sentences with all sentence-based style transfer techniques.}
    \label{tab:results-semantics-METEOR}
  \end{center}
\end{table}

\subsubsection{Semantic retainment}
\label{sec:results-semantic}

METEOR and manual evaluation scores are presented in Tables \ref{tab:results-semantics-METEOR}--\ref{tab:results-semantics-manual}, and user study results in Table \ref{tab:results-user-study}.

\noindent{\textbf{METEOR:}}
\sysname and A$^4$NT always clearly outperformed CAE and BT in METEOR. Especially on the smaller datasets (AB, TO), CAE and BT attained very poor scores ($<10$) that imply almost no semantic overlap with the original sentences.
A$^4$NT performed comparably to \sysname in the large datasets (YG, BA), but never exceeded \sysname-CNN, which was the highest performing \sysname-variant.
However, in the small datasets (AB, TO) A$^4$NT's scores dropped sharply.
Different \sysname-variants performed comparably.\footnote{\sysname-random performed the best overall, but we exclude it here because of its lack of targeted paraphrase selection. The METEOR scores of \sysname-random were $46.43$ (YG), $49.72$ (BA), $50.54$ (AB), and $49.88$ (TO).}
Unlike A$^4$NT, \sysname achieved high scores ($\sim 50$) on \textit{both} large and small datasets.

We also compared METEOR scores with each \sysname-module applied alone in the \sysname-LSTM variant. All achieved scores between $50$ and $67$. Simple rules remained the highest, as expected due to the small extent of paraphrases they produce. Typos had the largest range of variation ($50-67$), which demonstrates their dependency on the extent to which possible typos are available in the target class training data (Section \ref{sec:typos}). Grammatical transformations, PPDB, and WordNet performed similarly (in the range $55-63$), WordNet being systematically slightly higher than the rest. A likely reason for this is METEOR's bias toward WordNet synonyms as opposed to the kinds of paraphrases produced with grammatical transformations or PPDB. In contrast, in manual evaluation PPDB fared better than WordNet.

\noindent{\textbf{Manual evaluation:}}
Table \ref{tab:results-semantics-manual} presents our manual evaluation on $100$ sentences from each two-class dataset.
For practical reasons we limited our manual \sysname-evaluation to only one variant. We chose \sysname-LR$_\text{s}$ for two reasons. First, compared to other variants, it had neither the highest nor lowest overall METEOR score (Table \ref{tab:results-semantics-METEOR}), which indicates that the manual evaluations are not likely to either over- or underestimate the general performance of \sysname. Second, it implements the most realistic access assumptions out of all (non-random) \sysname-variants (Section \ref{sec:problem-statement}, Table \ref{tab:assumptions}).

\begin{table}[t!]
  \begin{center}
    \begin{tabular}{l|cccc}
     Technique & YG & BA & AB & TO \\ \hline
     CAE & $\textnormal{2\%}$ & $\textnormal{15\%}$ & $\textnormal{1\%}$ & $\textnormal{0\%}$ \\
     BT & $\textnormal{3\%}$ & $\textnormal{3\%}$ & $\textnormal{0\%}$ & $\textnormal{0\%}$ \\
     A$^4$NT & $\textnormal{31\%}$ & $\textnormal{44\%}$ & $\textnormal{16\%}$ & $\textnormal{6\%}$ \\
     \sysname-LR$_\text{s}$ & $\fbox{\textnormal{54\%}}$ & $\fbox{\textnormal{59\%}}$ & $\fbox{\textnormal{60\%}}$ & $\fbox{\textnormal{61\%}}$ \\
    \end{tabular}
    \caption{Manual evaluation: rate of acceptable paraphrases from $100$ sentences in each dataset ($50$ to both directions), transformed with each sentence-based technique.}
    \label{tab:results-semantics-manual}
  \end{center}
\end{table}

\begin{table}[t!]
  \begin{center}
    \begin{tabular}{l|cccc}
    Technique & Mean & Median & $\geq4$ & $5$ \\ \hline
    CAE & $\textnormal{0.8}$ & $\textnormal{0}$ & $\textnormal{5\%}$ & $\textnormal{2\%}$ \\
    BT & $\textnormal{0.9}$ & $\textnormal{0}$ & $\textnormal{8\%}$ & $\textnormal{3\%}$ \\
    A$^4$NT & $\textnormal{1.7}$ & $\textnormal{1}$ & $\textnormal{20\%}$ & $\textnormal{9\%}$ \\
    \sysname-LR$_\text{s}$ & $\textnormal{\fbox{2.7}}$ & $\textnormal{\fbox{3}}$ & $\textnormal{\fbox{41\%}}$ & $\textnormal{\fbox{24\%}}$ \\
    \end{tabular}
    \caption{User study results: grade statistics from human evaluations of meaning similarity from $400$ sentences from YG, transformed with each sentence-based technique (grade scale $0-5$).}
    \label{tab:results-user-study}
  \end{center}
\end{table}

\begin{table*}[t!]
  \begin{center}
    \begin{tabular}{l|ccc|ccc|ccc|ccc}
    \multirow{2}{*}{Technique} & \multicolumn{3}{c|}{YG} & \multicolumn{3}{c|}{BA} & \multicolumn{3}{c|}{AB} & \multicolumn{3}{c}{TO} \\
    & LSTM & CNN & WP & LSTM & CNN & WP & LSTM & CNN & WP & LSTM & CNN & WP \\ \hline

    CAE & $\textnormal{0.31}$ & $\textnormal{0.30}$ & $\textnormal{0.21}$ & $\textnormal{0.15}$ & $\textnormal{0.16}$ & $\textnormal{0.13}$ & $\textnormal{0.74}$ & $\textnormal{0.77}$ & $\textnormal{0.55}$ & $\textnormal{0.34}$ & $\textnormal{0.20}$ & $\textnormal{0.28}$ \\
    
    BT & $\textnormal{0.33}$ & $\textnormal{0.34}$ & $\textnormal{0.20}$ & $\textnormal{0.15}$ & $\textnormal{0.17}$ & $\textnormal{0.09}$ & $\fbox{\textnormal{0.77}}$ & $\fbox{\textnormal{0.84}}$ & $\textnormal{0.59}$ & $\textnormal{0.49}$ & $\fbox{\textnormal{0.33}}$ & $\textnormal{0.27}$ \\
    
    A$^4$NT & $\textnormal{0.16}$ & $\textnormal{0.17}$ & $\textnormal{0.10}$ & $\textnormal{0.10}$ & $\textnormal{0.11}$ & $\textnormal{0.05}$ & $\textnormal{0.19}$ & $\textnormal{0.22}$ & $\textnormal{0.14}$ & $\fbox{\textnormal{0.53}}$ & $\textnormal{0.07}$ & $\textnormal{0.33}$ \\

    \sysname-CNN & $\textnormal{0.39}$ & $\fbox{\textnormal{0.49}}$ & $\textnormal{0.19}$ & $\textnormal{0.21}$ & $\fbox{\textnormal{0.35}}$ & $\textnormal{0.08}$ & $\textnormal{0.26}$ & $\textnormal{0.39}$ & $\textnormal{0.16}$ & $\textnormal{0.13}$ & $\textnormal{0.21}$ & $\textnormal{0.12}$ \\

    \sysname-LSTM & $\fbox{\textnormal{0.44}}$ & $\textnormal{0.40}$ & $\textnormal{0.22}$ & $\fbox{\textnormal{0.37}}$ & $\textnormal{0.27}$ & $\textnormal{0.11}$ & $\textnormal{0.47}$ & $\textnormal{0.32}$ & $\textnormal{0.33}$ & $\textnormal{0.47}$ & $\textnormal{0.09}$ & $\textnormal{0.17}$ \\

    \sysname-WP & $\textnormal{0.26}$ & $\textnormal{0.23}$ & $\fbox{\textnormal{0.48}}$ & $\textnormal{0.16}$ & $\textnormal{0.17}$ & $\fbox{\textnormal{0.34}}$ & $\textnormal{0.21}$ & $\textnormal{0.22}$ & $\fbox{\textnormal{0.60}}$ & $\textnormal{0.15}$ & $\textnormal{0.06}$ & $\fbox{\textnormal{0.59}}$ \\

    \sysname-LR$_\text{d}$ & $\textnormal{0.39}$ & $\textnormal{0.40}$ & $\textnormal{0.16}$ & $\textnormal{0.28}$ & $\textnormal{0.29}$ & $\textnormal{0.11}$ & $\textnormal{0.26}$ & $\textnormal{0.25}$ & $\textnormal{0.29}$ & $\textnormal{0.32}$ & $\textnormal{0.09}$ & $\textnormal{0.17}$ \\

    \sysname-LR$_\text{s}$ & $\textnormal{0.36}$ & $\textnormal{0.36}$ & $\textnormal{0.15}$ & $\textnormal{0.22}$ & $\textnormal{0.21}$ & $\textnormal{0.09}$ & $\textnormal{0.21}$ & $\textnormal{0.22}$ & $\textnormal{0.20}$ & $\textnormal{0.23}$ & $\textnormal{0.07}$ & $\textnormal{0.14}$ \\

    \sysname-random & $\textnormal{0.12}$ & $\textnormal{0.12}$ & $\textnormal{0.08}$ & $\textnormal{0.04}$ & $\textnormal{0.05}$ & $\textnormal{0.01}$ & $\textnormal{0.10}$ & $\textnormal{0.11}$ & $\textnormal{0.09}$ & $\textnormal{0.09}$ & $\textnormal{0.03}$ & $\textnormal{0.09}$ \\

    \end{tabular}
    \captionsetup{justification=centering}
    \caption{Author profiler accuracy decrease in sentence-based datasets, best (highest) scores framed.}
    \label{tab:results-imitation}
  \end{center}
\end{table*}

CAE and BT produced practically \textit{no acceptable paraphrases}. This was especially true in the small datasets (AB, TO), where imitations bore no resemblance to the original sentence and simply repeated certain words prevalent in the target corpus.

With A$^4$NT, sentence \textit{reproduction} was much more common than anywhere else, but actual paraphrases remained rare.
For example, A$^4$NT's acceptable paraphrase rate in BA decreases from $44\%$ to only $2\%$ when reproductions are excluded.
A$^4$NT also generated a large number of \textit{omissions}, reproducing only part of the original sentence without any changes or additions.
Such pure omissions were rare with other techniques. These characteristics are likely due to A$^4$NT's training function, which includes a \textit{reconstruction loss} \cite{Shettyetal2018} (Section \ref{sec:encoder-decoder-baselines}).
In the small datasets (AB, TO) A$^4$NT's semantic retainment starkly declined to almost non-existent.

\sysname clearly stood out by producing many acceptable paraphrases.
In contrast to the baselines, \sysname's performance was similar across all four datasets.
The most prevalent \sysname-transformation was paraphrase replacement from PPDB or WordNet, most commonly targeting a single word.
Typos and grammatical transformations were rare in the manually evaluated test sets, but some were encountered.
When PPDB- and WordNet-replacement could be distinguished, PPDB-replacement fared better in semantic retainment. Most \sysname-errors were caused by contextually unsuitable WordNet synonym choice (e.g. \textit{man}$\rightarrow$\textit{mankind}). Such errors are due to faulty word sense disambiguation (Section \ref{sec:paraphrase-replacement}), improving which is an important challenge for future research.

\noindent{\textbf{User study:}}
Table \ref{tab:results-user-study} presents the user study results.
\sysname clearly outperformed all baselines.
As expected, CAE and BT performed especially poorly.
The majority of \sysname-imitations were on the upper half of the six-point scale ($3-5$), whereas the majority of all baseline imitations had the lowest grades ($0-1$).

\subsubsection{Style transfer}
\label{sec:results-imitation}

We present style transfer results on two-class settings and multi-class author imitation.

\noindent{\textbf{Two-class tests:}}
To evaluate style transfer success, we measured \textit{accuracy decrease}: i.e. how much original accuracy dropped after style transfer.
Table \ref{tab:results-imitation} provides these results.
Table \ref{tab:results-twoclass-orig} shows profiling accuracies on original test sets with the three profilers (Section \ref{sec:eval_imitation_success}).

A$^4$NT's performance was the weakest everywhere except TO.
CAE and BT achieved almost full imitation in AB and the Obama$\rightarrow$Trump direction of TO.
This was unsurprising since they simply repeated words unrelated to the original (Section \ref{sec:results-semantic}).
However, both showed a clear bias toward the Trump class, and failed in the Trump$\rightarrow$Obama direction.

\begin{table}[t!]
  \begin{center}
    \begin{tabular}{l|cccc}
    Profiler & YG & BA & AB & TO \\ \hline
    LSTM & $\textnormal{0.83}$ & $\textnormal{0.62}$ & $\textnormal{0.91}$ & $\textnormal{0.82}$ \\
    CNN & $\textnormal{0.82}$ & $\textnormal{0.63}$ & $\textnormal{0.93}$ & $\textnormal{0.64}$ \\
    WP & $\textnormal{0.75}$ & $\textnormal{0.59}$ & $\textnormal{0.82}$ & $\textnormal{0.74}$ \\
    \end{tabular}
    \caption{Original author profiling accuracies.}
    \label{tab:results-twoclass-orig}
  \end{center}
\end{table}

In large datasets (YG, BA), \textit{all} (non-random) variants of \sysname \textit{outperformed all baselines}. This happened even under the weakest access assumptions, i.e. \sysname-LR$_\text{s}$.
In small datasets (AB, TO), \sysname retained similar performance but baselines increased theirs.
\sysname-LR$_\text{d}$ achieved $5\%$ better average accuracy decrease than \sysname-LR$_\text{s}$, and query access to the author profiler (CNN or LSTM) increased it $10\%-20\%$.
Query access to WP allowed effective black-box evasion of WP, but did not reach the performance of other variants with other profilers.
\sysname-random expectedly took accuracy down the least.

\begin{table}[t!]
  \begin{center}
    \begin{tabular}{l|cccccc}
    Technique & YG & BA & AB & TO & Avg. \\ \hline
    Grammatical & $\textnormal{0.12}$ & $\textnormal{0.07}$ & $\textnormal{0.14}$ & $\textnormal{0.08}$ & $\textnormal{0.10}$ \\
    Simple rules & $\textnormal{0.01}$ & $\textnormal{0.04}$ & $\textnormal{0.03}$ & $\textnormal{0.05}$ & $\textnormal{0.03}$ \\
    PPDB & $\fbox{\textnormal{0.19}}$ & $\textnormal{0.18}$ & $\textnormal{0.13}$ & $\fbox{\textnormal{0.25}}$ & $\fbox{\textnormal{0.19}}$ \\
    WordNet & $\textnormal{0.15}$ & $\fbox{\textnormal{0.20}}$ & $\textnormal{0.13}$ & $\textnormal{0.23}$ & $\textnormal{0.18}$ \\
    Typos & $\textnormal{0.02}$ & $\textnormal{0.10}$ & $\fbox{\textnormal{0.15}}$ & $\textnormal{0.08}$ & $\textnormal{0.09}$ \\ \hline
    All & $\textnormal{0.44}$ & $\textnormal{0.37}$ & $\textnormal{0.47}$ & $\textnormal{0.47}$ & $\textnormal{0.46}$ \\
    \end{tabular}
    \caption{Profiler accuracy decrease with \sysname modules individually and together (LSTM profiler, \sysname-LSTM variant).}
    \label{tab:results-parchoice-modules}
  \end{center}
\end{table}

\begin{table}[t!]
  \begin{center}
    \begin{tabular}{l|cccccc}
    & Gram. & Simple & PPDB & WordNet & Typos \\ \hline
    Gram. & $\textnormal{100\%}$ & $\textnormal{93\%}$ & $\textnormal{80\%}$ & $\textnormal{80\%}$ & $\textnormal{89\%}$ \\
    Simple & $\textnormal{93\%}$ & $\textnormal{100\%}$ & $\textnormal{83\%}$ & $\textnormal{84\%}$ & $\textnormal{95\%}$ \\
    PPDB & $\textnormal{80\%}$ & $\textnormal{83\%}$ & $\textnormal{100\%}$ & $\textnormal{81\%}$ & $\textnormal{82\%}$ \\
    WordNet & $\textnormal{80\%}$ & $\textnormal{84\%}$ & $\textnormal{81\%}$ & $\textnormal{100\%}$ & $\textnormal{83\%}$ \\
    Typos & $\textnormal{89\%}$ & $\textnormal{95\%}$ & $\textnormal{82\%}$ & $\textnormal{83\%}$ & $\textnormal{100\%}$ \\
    \end{tabular}
    \caption{Prediction overlap between \sysname modules (combined from all datasets, LSTM profiler, \sysname-LSTM variant).}
    \label{tab:results-parchoice-modules-overlap}
  \end{center}
\end{table}

We additionally applied each \sysname module separately on the \sysname-LSTM variant, and compared accuracy decrease (Table \ref{tab:results-parchoice-modules}) and prediction overlap (Table \ref{tab:results-parchoice-modules-overlap}) between the modules on the LSTM profiler.
All techniques had at least a minor impact.
PPDB and WordNet were the most effective overall and had similar success ($13\%-20\%$).
Grammatical transformations outperformed them in AB, but were less successful otherwise.
Typos had the largest variation, ranging from almost nonexistent in YG ($2\%$) to being the most effective in AB ($15\%$).
Simple rules were expectedly the least effective when applied alone ($1\%-5\%$).
The most effective techniques (PPDB and WordNet) were also the most diverse, having the least prediction overlap with other techniques ($80\%-84\%$) and each other ($81\%$).

\noindent{\textbf{Proofreading:}}
To emulate a scenario where the author manually checks the results of style transfer, we proofread $50$ transformed sentences from both the \textit{male} and \textit{female} test sets of YG.
For CAE and BT proofreading often required reproducing the original sentence due to very poor semantic retainment (Section \ref{sec:results-semantic}).
Proofreading negatively impacted style transfer with all techniques, but markedly less with \sysname than the baselines (Table \ref{tab:results-proofreading}).

\begin{table}[b!]
  \begin{center}
    \begin{tabular}{l|cccc}
    
    Profiler & CAE & BT & A$^4$NT & \sysname-LR$_s$ \\ \hline
    LSTM & $\textnormal{0.11}$ &
    $\textnormal{0.14}$ &
    $\textnormal{0.09}$ &
    $\fbox{\textnormal{0.31}}$ \\

    CNN & $\textnormal{0.12}$ &
    $\textnormal{0.11}$ &
    $\textnormal{0.09}$ &
    $\fbox{\textnormal{0.26}}$ \\

    WP & $\textnormal{0.09}$ &
    $\textnormal{0.09}$ &
    $\textnormal{0.08}$ &
    $\fbox{\textnormal{0.10}}$ \\

    \end{tabular}
    \caption{Profiler accuracy decrease on a YG-subset after manual proofreading; best (highest) scores framed.}
    \label{tab:results-proofreading}
  \end{center}
\end{table}

\begin{table}[t!]
  \begin{center}
    \begin{tabular}{c|c|ccc}

    \multirow{2}{*}{Technique} & \multirow{2}{*}{Profiler} & \multirow{2}{*}{\shortstack[c]{Source \\ decrease}} & \multirow{2}{*}{\shortstack[c]{Target \\ increase}} & \multirow{2}{*}{Imitation} \\ &&& \\ \hline

    \multirow{2}{*}{\sysname-LSTM} & LSTM & $\textnormal{0.39}$ & $\fbox{\textnormal{0.29}}$ & $\fbox{\textnormal{15/20}}$ \\
    & CNN & $\textnormal{0.28}$ & $\textnormal{0.20}$ & $\textnormal{1/20}$ \\ \hline

    \multirow{2}{*}{\sysname-CNN} & LSTM & $\textnormal{0.21}$ & $\textnormal{0.13}$ & $\textnormal{0/20}$ \\
    & CNN & $\fbox{\textnormal{0.40}}$ & $\textnormal{0.27}$ & $\textnormal{7/20}$ \\

    \end{tabular}
    \caption{Five-class author imitation in the blog author dataset with \sysname (query access to LSTM/CNN): average source class accuracy decrease, target class accuracy increase, and imitation success (imitation threshold: majority of source class documents assigned to target class).}
    \label{tab:results-multiclass}
  \end{center}
\end{table}

\noindent{\textbf{Multi-class tests:}}
We evaluated five-class author imitation to every direction on a blog author dataset (Section \ref{sec:datasets}).
Since \sysname-CNN and \sysname-LSTM had the best overall performance in two-class settings, we experimented on these variants on the CNN and LSTM profilers. We discarded WP here because it failed to achieve high profiling accuracies on the original five-class test sets.
On each author's test set, we considered both the \textit{accuracy decrease of the original class}, and the \textit{accuracy increase of the target class}.
Additionally, if the majority of the source author test sentences was assigned to the target author, we considered imitation to succeed for that source-target pair.
Table \ref{tab:results-multiclass} summarizes the results from all $20$ imitation directions. \sysname-LSTM's overall performance was superior to \sysname-CNN's.
\sysname-LSTM reached the threshold $75\%$ of the time on the LSTM profiler.

\subsection{Document-level experiments}
\label{sec:document-based}

We compare \sysname to the rule-based PAN2016 \cite{Karadzhovetal2017} and Mutant-X \cite{Mahmoodetal2019} baselines (Section \ref{sec:rule-based-baselines}) in semantic retainment (\ref{sec:document-semantic}) and style transfer (\ref{sec:document-style-transfer}).
\sysname is markedly better in semantic retainment than either baseline, but Mutant-X remains stronger in style transfer.
A combination of \sysname and Mutant-X outperforms prior rule-based schemes in both style transfer and semantic retainment.

\subsubsection{Semantic retainment}
\label{sec:document-semantic}

We use three measures of document-level semantic retainment: (i) METEOR score, (ii) rate of sentences transformed per document, and (iii) manual evaluation of paraphrases in \textit{transformed} sentences.
Originally misclassified documents were discarded from the evaluation, since they were left unchanged.

\begin{table}[h!]
  \begin{center}
    \begin{tabular}{l|cccc}
    \multirow{2}{*}{Technique} & \multicolumn{4}{c}{METEOR} \\
    & BG & EBG$_5$ & EBG$_{10}$ & EBG$_{45}$ \\ \hline
    PAN2016 & $\textnormal{43.17}$ & $\textnormal{45.93}$ & $\textnormal{46.96}$ & $\textnormal{47.15}$ \\
    Mutant-X & $\textnormal{56.71}$ & $\textnormal{57.23}$ & $\textnormal{53.46}$ & $\textnormal{56.75}$ \\
    \sysname & $\fbox{\textnormal{77.40}}$ & $\fbox{\textnormal{65.41}}$ & $\fbox{\textnormal{66.26}}$ & $\fbox{\textnormal{69.34}}$ \\
    \sysname{}+Mutant-X & $\textnormal{61.61}$ & $\textnormal{60.25}$ & $\textnormal{62.31}$ & $\textnormal{65.86}$ \\
    \end{tabular}
    \caption{METEOR scores between original and transformed documents from rule-based techniques (originally misclassified documents discarded).}
    \label{tab:rule-METEOR}
  \end{center}
\end{table}

\noindent{\textbf{METEOR:}}
All techniques achieved relatively high METEOR scores (Table \ref{tab:rule-METEOR}), with \sysname outperforming PAN2016 (by $19-34$ points) and Mutant-X (by $8-21$ points).
Combining \sysname with Mutant-X also always retained a higher score than Mutant-X alone (by $9-3$ points).

\noindent{\textbf{Transformed sentence rates:}}
PAN2016 transformed almost all sentences per document, Mutant-X a clear majority ($>70\%$), and \sysname only $6\%-25\%$ (Table \ref{tab:document-transformed-rates}).
This is a major divergence among the techniques, as it makes \sysname much more likely to retain semantics by focusing transformations only on a minority of sentences.
While the transformed sentence rate increased when Mutant-X was applied after \sysname, it remained significantly lower than with Mutant-X alone.

\begin{table}[t!]
  \begin{center}
    \begin{tabular}{l|cccc}
    \multirow{2}{*}{Technique} & \multicolumn{4}{c}{Transformed sentences} \\
    & BG & EBG$_5$ & EBG$_{10}$ & EBG$_{45}$ \\ \hline
    PAN2016 & $\textnormal{98\%}$ & $\textnormal{98\%}$ & $\textnormal{98\%}$ & $\textnormal{96\%}$ \\
    Mutant-X & $\textnormal{71\%}$ & $\textnormal{72\%}$ & $\textnormal{79\%}$ & $\textnormal{73\%}$ \\
    \sysname & $\fbox{\textnormal{6\%}}$ & $\fbox{\textnormal{25\%}}$ & $\fbox{\textnormal{11\%}}$ & $\fbox{\textnormal{7\%}}$ \\
    \sysname{}+Mutant-X & $\textnormal{43\%}$ & $\textnormal{41\%}$ & $\textnormal{22\%}$ & $\textnormal{14\%}$ \\
    \end{tabular}
    \caption{Average transformed sentence rates per document (originally misclassified documents discarded).}
    \label{tab:document-transformed-rates}
  \end{center}
\end{table}

\noindent{\textbf{Manual evaluation:}}
To ensure fair comparison, we manually evaluated each technique on the \textit{same} sentences. Across all four datasets there were $81$ sentences that all techniques had made changes to. Table \ref{tab:document-manual} shows manual evaluation results on these sentences.

We provide an error analysis, categorizing errors to four types based on decreasing severity:

\begin{itemize}
\item{} \textbf{Antonym}: opposite meaning
\item{} \textbf{Word}: non-antonym but different meaning
\item{} \textbf{Context}: possible paraphrase but wrong context
\item{} \textbf{Grammar}: correct paraphrase but wrong grammar
\end{itemize}

\noindent{}Mutant-X produced $19$ antonyms across all $81$ sentences, including e.g. \textit{more}$\rightarrow$\textit{less} and \textit{easier}$\rightarrow$\textit{harder}. While Mutant-X uses sentiment-based neighbour ranking \cite{Yuetal2018} to avoid antonyms (Section \ref{sec:rule-based-baselines}), sentiment is only one possible source of antonymity.
In contrast, antonyms were absent in PAN2016, and only one was found in \sysname: \textit{defended}$\rightarrow$\textit{opposed}. The most likely source for this error was PPDB.
Other word errors were the most common in all techniques, but much rarer in \sysname than elsewhere.
In \sysname the rate of context errors was the same as in PAN2016, but their percentage of all errors larger.
These included e.g. \textit{issue}$\rightarrow$\textit{number}, which would be correct in a magazine-related context but not otherwise.
All techniques produced a few ($1-4$) purely grammatical errors, mostly due to inflection.

Overall, Mutant-X made markedly more semantic errors than either PAN2016 or \sysname.
This result seems to contrast PAN2016 having a lower METEOR score. This may be due to PAN2016 adding words (such as \textit{Additionally}) before sentences, which usually does not negatively impact semantics but brings n-gram overlap (and hence METEOR) down.
Applying \sysname before Mutant-X significantly reduced the most fatal errors (antonym or word error) produced by Mutant-X.

\begin{table}[t!]
  \begin{center}
    \begin{tabular}{l|c|cccc}
    \multirow{2}{*}{Technique} & \multirow{2}{*}{\shortstack{Acceptable \\ Paraphrases}} & \multicolumn{4}{c}{Error count} \\
    && A & W & C & G \\ \hline
    PAN2016 & $\textnormal{44\%}$ & $\textnormal{0}$ & $\textnormal{32}$ & $\textnormal{12}$ & $\textnormal{1}$ \\
    Mutant-X & $\textnormal{32\%}$ & $\textnormal{19}$ & $\textnormal{62}$ & $\textnormal{7}$ & $\textnormal{4}$\\
    \sysname & $\fbox{\textnormal{60\%}}$ & $\textnormal{1}$ & $\textnormal{20}$ & $\textnormal{12}$ & $\textnormal{2}$ \\
    \sysname{}+Mutant-X & $\textnormal{53\%}$ & $\textnormal{2}$ & $\textnormal{40}$ & $\textnormal{16}$ & $\textnormal{1}$ \\
    \end{tabular}
    \caption{Manual evaluation results from rule-based techniques (calculated from $81$ sentences transformed by all techniques). \\
    Error types: A=antonym; W=word; C=context; G=grammar.}
    \label{tab:document-manual}
  \end{center}
\end{table}

\begin{table}[t!]
  \begin{center}
    \begin{tabular}{l|cccc}
    \multirow{2}{*}{Technique} & \multicolumn{4}{c}{Corpus} \\
    & BG & EBG$_5$ & EBG$_{10}$ & EBG$_{45}$ \\ \hline
    PAN2016 & $\textnormal{50\%}$ & $\textnormal{13\%}$ & $\textnormal{32\%}$ & $\textnormal{76\%}$ \\
    Mutant-X & $\fbox{\textnormal{82\%}}$ & $\textnormal{60\%}$ & $\textnormal{87\%}$ & $\textnormal{96\%}$ \\
    \sysname & $\textnormal{36\%}$ & $\textnormal{47\%}$ & $\textnormal{84\%}$ & $\textnormal{91\%}$ \\
    \sysname{+}Mutant-X & $\fbox{\textnormal{82\%}}$ & $\fbox{\textnormal{77\%}}$ & $\fbox{\textnormal{97\%}}$ & $\fbox{\textnormal{100\%}}$ \\
    \end{tabular}
    \caption{Successful style transfer rates of rule-based techniques (originally misclassified documents discarded).}
    \label{tab:rule-style-transfer}
  \end{center}
\end{table}

\subsubsection{Style transfer}
\label{sec:document-style-transfer}

Table \ref{tab:rule-style-transfer} summarizes document-level style transfer performance.
We define the \textit{successful style transfer rate} as the frequency at which an originally correctly classified document was incorrectly classified after style transfer.
Mutant-X had superior performance to PAN2016 and \sysname across all datasets, although the difference to \sysname was only minor in EBG$_5$ and EBG$_{10}$ ($3\%-5\%$).
However, when Mutant-X was applied after \sysname, we reached the highest result on all datasets.
Given that this combination also achieved significantly higher semantic retainment than Mutant-X alone (Section \ref{sec:document-semantic}), it constitutes the state-of-the-art solution to document-level style transfer on these datasets.

However, we note a problem in assuming \textit{query access} to the adversary's random forest profiler.
The profiler involves randomness in training, and \textit{re-training} it on the same training data mostly \textit{undoes} the effect of style transfer on query-access techniques.
Table \ref{tab:results-ebg-new-profiler} shows this effect on EBG$_5$.
As PAN2016 does not use query access, it is not vulnerable to this problem.
Query access can thus result in highly \textit{profiler-specific} transformations, which are \textit{non-transferable} across profilers.

\begin{table}[t!]
  \begin{center}
  \begin{tabular}{l|cc}
    \multirow{2}{*}{Technique} & \multirow{2}{*}{\shortstack{Original \\ profiler}} & \multirow{2}{*}{\shortstack{New \\ profiler}} \\ \\ \hline
    \textnormal{PAN2016} & $\textnormal{13\%}$ & $\textnormal{20\%}$ \\
    \textnormal{Mutant-X} & $\textnormal{60\%}$ & $\textnormal{17\%}$ \\
    \textnormal{\sysname} & $\textnormal{47\%}$ & $\textnormal{20\%}$ \\
    \end{tabular}
    \caption{Style transfer success rates on EBG$_5$ with the original (queried) profiler and a new re-trained random forest profiler.}
    
    \label{tab:results-ebg-new-profiler}
  \end{center}
\end{table}

Problematically, most style transfer research has so far relied on query access. In particular, with the exception of PAN2016 \cite{Karadzhovetal2017}, all our baseline techniques were originally applied in query access settings \cite{Shenetal2017, Prabhumoyeetal2018, Shettyetal2018, Mahmoodetal2019}.
Since query access is not a realistic requirement for the author (Section \ref{sec:problem-statement}), and given its lack of transferability to different profilers, we recommend moving beyond query access in style transfer research.
While we were able to do so successfully on sentence-based data (Section \ref{sec:results-sentence-based}), on the document-level it remains a challenge for future work.

\subsection{Thwarting style transfer via adversarial training}
\label{sec:results-adversarial-training}

To test the effectiveness of adversarial training for countering style transfer, we applied it to every technique on the AB dataset with the LSTM profiler. We chose this setting because it achieved a high original test set accuracy as well as the highest overall accuracy change (Section \ref{sec:results-imitation}, Table \ref{tab:results-imitation}).
Table \ref{tab:results-adversarial-training} presents the results.

Profiling accuracies on original test sets remained high, but were taken down in all cases except two (A$^4$NT and \sysname-LSTM in the Alice$\rightarrow$Bob direction, which increased by $2\%$). Accuracy on the transformed test set increased drastically with most techniques, and mostly undid the effect of style transfer.\footnote{Apart from \sysname-random (discussed separately), the only exception was A$^4$NT in the Bob$\rightarrow$Alice direction, where profiling accuracy decreased by $6\%$ in transformed sentences. Since A$^4$NT's original performance was not strong on AB to begin with, and original profiling accuracy also decreased in the Bob$\rightarrow$Alice direction (by $9\%$), we do not take this exception to affect the overall conclusion.}
The important exception was \sysname-random, which we discuss below.

Our results illustrate a major problem in style transfer techniques that rely on the author having query access to the adversary's profiler, or a surrogate profiler that accurately approximates its performance. Crucially, such access assumptions go \textit{both ways}: if the author can access/approximate the adversary's profiler locally, the adversary can also access/approximate the author's local profiler for adversarial training. This problem is fundamental to all of our techniques \textit{except \sysname-random}, which performs paraphrase selection independently of the profiler.

\begin{table}[t!]
  \begin{center}
  \begin{tabular}{cccc}
  
    Technique & Direction & Original & Transformed \\ \hline
    
    \multirow{2}{*}{CAE} & A$\rightarrow$B & $\textnormal{0.88} \rightarrow \textnormal{0.78}$ & $\textnormal{0.18} \rightarrow \textnormal{0.90}$ \\
    & B$\rightarrow$A & $\textnormal{0.94} \rightarrow \textnormal{0.79}$ & $\textnormal{0.16} \rightarrow \textnormal{0.87}$ \\ \hline

    \multirow{2}{*}{BT} & A$\rightarrow$B & $\textnormal{0.88} \rightarrow \textnormal{0.85}$ & $\textnormal{0.18} \rightarrow \textnormal{0.78}$ \\
    & B$\rightarrow$A & $\textnormal{0.94} \rightarrow \textnormal{0.88}$ & $\textnormal{0.11} \rightarrow \textnormal{0.64}$ \\ \hline

    \multirow{2}{*}{A$^4$NT} & A$\rightarrow$B & $\textnormal{0.88} \rightarrow \textnormal{0.90}$ & $\textnormal{0.74} \rightarrow \textnormal{0.86}$ \\
    & B$\rightarrow$A & $\textnormal{0.94} \rightarrow \textnormal{0.85}$ & $\textnormal{0.70} \rightarrow \textnormal{0.64}$ \\ \hline

    \multirow{2}{*}{\sysname-LSTM} & A$\rightarrow$B & $\textnormal{0.88} \rightarrow \textnormal{0.90}$ & $\textnormal{0.36} \rightarrow \textnormal{0.88}$ \\
    & B$\rightarrow$A & $\textnormal{0.94} \rightarrow \textnormal{0.88}$ & $\textnormal{0.52} \rightarrow \textnormal{0.92}$ \\ \hline

    \multirow{2}{*}{\sysname-random} & A$\rightarrow$B & $\textnormal{0.88} \rightarrow \textnormal{0.79}$ & $\textnormal{0.74} \rightarrow \textnormal{0.47}$ \\
    & B$\rightarrow$A & $\textnormal{0.94} \rightarrow \textnormal{0.72}$ & $\textnormal{0.89} \rightarrow \textnormal{0.38}$ \\

    \end{tabular}
    \caption{LSTM author profiler accuracy without$\rightarrow$with adversarial training, on both original and transformed test sets of AB.}
    \label{tab:results-adversarial-training}
  \end{center}
\end{table}

In striking contrast to other techniques, adversarial training with \sysname-random reduced profiling accuracy on \textit{both} original and transformed sentences. Hence, while \sysname-random expectedly performed the least effectively in style transfer (Section \ref{sec:results-imitation}), it was the only technique that could effectively resist adversarial training as a counter-measure.
One possible reason for this is the large range of stylistic variants produced by \sysname's paraphrase generation stage, which results in author-specific stylistic markers becoming less prominent in the adversarial training set.

Adversarial training is also expected to be more challenging when the number of classes is increased.
We produced \sysname-LSTM imitations with random target classes from the five-class training set, and appended them to the original five-class training data. We trained the LSTM profiler on this adversarial training data, and tested it against \sysname-LSTM imitations with randomly selected targets. Results are shown in Table \ref{tab:results-adversarial-training-multiclass}. As predicted, the effectiveness of adversarial training was reduced, although still clearly present.
However, the adversary could adopt a two-step profiling procedure of first detecting the imitation target by classifying the transformed document, and then producing the adversarial training set using this target.

\begin{table}[t!]
  \begin{center}
  \begin{tabular}{ccc}
    Author & Original & Transformed \\ \hline
    A1 & $\textnormal{0.75} \rightarrow \textnormal{0.73}$ & $\textnormal{0.27} \rightarrow \textnormal{0.61}$ \\
    A2 & $\textnormal{0.81} \rightarrow \textnormal{0.76}$ & $\textnormal{0.44} \rightarrow \textnormal{0.71}$ \\
    A3 & $\textnormal{0.64} \rightarrow \textnormal{0.60}$ & $\textnormal{0.26} \rightarrow \textnormal{0.50}$ \\
    A4 & $\textnormal{0.71} \rightarrow \textnormal{0.74}$ & $\textnormal{0.29} \rightarrow \textnormal{0.66}$ \\
    A5 & $\textnormal{0.64} \rightarrow \textnormal{0.72}$ & $\textnormal{0.35} \rightarrow \textnormal{0.65}$ \\
    \end{tabular}
    \caption{LSTM author profiler accuracy without$\rightarrow$with adversarial training, on both original and transformed five-class test sets with randomized target authors on \sysname-LSTM.}
    \label{tab:results-adversarial-training-multiclass}
  \end{center}
\end{table}

On the document-level, the effect of adversarial training is difficult to evaluate due to the techniques' brittleness to re-training the profiler. While we were able to bring style transfer performance down on the EBG$_5$ dataset via adversarial training, this also happened with regular re-training (Table \ref{tab:results-ebg-new-profiler}). As discussed in Section \ref{sec:document-style-transfer}, we believe this is due to the high profiler-specificity of techniques relying on query access on the document-level, which remains an important problem for future research.
\section{Conclusions and future work}
\label{sec:conclusion}

We presented \sysname: a novel style transfer technique based on combinatorial paraphrase generation and style-specific paraphrase selection.
\sysname considerably improves on the state-of-the-art in retaining semantic content.
In sentence-level experiments, it outperformed the only encoder-decoder baseline technique with competitive semantic retainment (A$^4$NT).
On the document-level, combining \sysname with the best-performing rule-based baseline (Mutant-X) increased state-of-the-art performance in both style transfer and semantic retainment.
We thus endorse \sysname as the most viable style transfer technique overall. On data where \sysname has limited coverage, we recommend combining it with word embedding -based replacement.

An important extension of the present work is applying style transfer to more complex profiling schemes. In particular, \textit{abstaining classifiers} \cite{Chow1970, Ferrietal2004} can be used to detect whether the probability of \textit{any} target class is too low for prediction to be justified.
They could be used to filter potential cases of style transfer, as these are less likely to give clear target class predictions. These datapoints could then be subjected to further scrutiny.
Abstaining classifiers have been demonstrated to be beneficial for stylometry, and for detecting manually transformed documents \cite{Stolermanetal2014}. Their effectiveness against automatic techniques remains to be studied.

Another major conclusion we draw is that style transfer needs to properly address possible countermeasures.
The main challenge is achieving strong style transfer performance whilst preventing the adversary from replicating it by adversarial training.
We demonstrated this to be a realistic concern, but propose that its effectiveness can partly be hindered via increased randomization in paraphrasing.
We will continue to explore this issue in future work.

\section*{Acknowledgements}
We thank Andrei Kazlouski and Sam Spilsbury for their help in implementation and running experiments, and Mika Juuti for valuable discussions related to the project.
Tommi Gröndahl was funded by the Helsinki Doctoral Education Network in Information and Communications Technology (HICT).

\bibliographystyle{plain}
\bibliography{parchoice_refs}

\begin{table*}[h!]
  \begin{center}
  \begin{tabular}{c|c|c|cccccccccc}
  
    \multicolumn{1}{c|}{\multirow{4}{*}{Data}} & \multirow{4}{*}{Direction} & \multirow{4}{*}{Author profiler} & \multicolumn{10}{c}{\multirow{2}{*}{Profiler accuracy for source author}} \\
    \multicolumn{1}{c|}{} &&& \multicolumn{10}{c}{} \\ \cline{4-13}
    \multicolumn{1}{c|}{} &&& \multirow{2}{*}{Original} & \multirow{2}{*}{CAE} & \multirow{2}{*}{BT} & \multicolumn{1}{c|}{\multirow{2}{*}{A$^4$NT}} & \multicolumn{6}{c}{\sysname} \\
    \multicolumn{1}{c|}{} &&&&&& \multicolumn{1}{c|}{} & CNN & LSTM & WP & LR$_\textit{d}$ & LR$_\textit{s}$ & \multicolumn{1}{c}{random} \\ \hline
    
    \multicolumn{1}{c|}{\multirow{6}{*}{YG}} & \multirow{3}{*}{f$\rightarrow$m} & LSTM & $\textnormal{0.83}$ & $\textnormal{0.55}$ & $\textnormal{0.54}$ & \multicolumn{1}{c|}{$\textnormal{0.64}$} &
    $\textnormal{0.41}$ & $\fbox{\textnormal{0.35}}$ & $\textnormal{0.54}$ & $\textnormal{0.43}$ & $\textnormal{0.46}$ & \multicolumn{1}{c}{$\textnormal{0.69}$} \\
    \multicolumn{1}{c|}{} && CNN & $\textnormal{0.75}$ & $\textnormal{0.44}$ & $\textnormal{0.43}$ & \multicolumn{1}{c|}{$\textnormal{0.57}$}
    & $\fbox{\textnormal{0.21}}$ & $\textnormal{0.28}$ & $\textnormal{0.46}$ & $\textnormal{0.28}$ & $\textnormal{0.32}$ & \multicolumn{1}{c}{$\textnormal{0.57}$} \\
    \multicolumn{1}{c|}{} && WP & $\textnormal{0.71}$ & $\textnormal{0.50}$ & $\textnormal{0.55}$ & \multicolumn{1}{c|}{$\textnormal{0.57}$}
    & $\textnormal{0.49}$ & $\textnormal{0.44}$ & $\fbox{\textnormal{0.21}}$ & $\textnormal{0.54}$ & $\textnormal{0.54}$ & \multicolumn{1}{c}{$\textnormal{0.60}$} \\ \cline{2-13}
    
    \multicolumn{1}{c|}{} & \multirow{3}{*}{m$\rightarrow$f} & LSTM & $\textnormal{0.82}$ & $\textnormal{0.48}$ & $\textnormal{0.45}$ & \multicolumn{1}{c|}{$\textnormal{0.70}$}
    & $\textnormal{0.46}$ & $\fbox{\textnormal{0.43}}$ & $\textnormal{0.59}$ & $\textnormal{0.45}$ & $\textnormal{0.48}$ & \multicolumn{1}{c}{$\textnormal{0.72}$} \\
    \multicolumn{1}{c|}{} && CNN & $\textnormal{0.88}$ & $\textnormal{0.59}$ & $\textnormal{0.53}$ & \multicolumn{1}{c|}{$\textnormal{0.73}$}
    & $\fbox{\textnormal{0.45}}$ & $\textnormal{0.56}$ & $\textnormal{0.71}$ & $\textnormal{0.56}$ & $\textnormal{0.59}$ & \multicolumn{1}{c}{$\textnormal{0.82}$} \\
    \multicolumn{1}{c|}{} && WP & $\textnormal{0.78}$ & $\textnormal{0.56}$ & $\textnormal{0.54}$ & \multicolumn{1}{c|}{$\textnormal{0.72}$}
    & $\textnormal{0.61}$ & $\textnormal{0.60}$ & $\fbox{\textnormal{0.32}}$ & $\textnormal{0.64}$ & $\textnormal{0.65}$ & \multicolumn{1}{c}{$\textnormal{0.74}$} \\ \hline

    \multicolumn{1}{c|}{\multirow{6}{*}{BA}} & \multirow{3}{*}{a$\rightarrow$t} & LSTM & $\textnormal{0.54}$ & $\textnormal{0.26}$ & $\textnormal{0.28}$ & \multicolumn{1}{c|}{$\textnormal{0.35}$} &
    $\textnormal{0.38}$ & $\fbox{\textnormal{0.16}}$ & $\textnormal{0.37}$ & $\textnormal{0.26}$ & $\textnormal{0.33}$ & \multicolumn{1}{c}{$\textnormal{0.52}$} \\
    \multicolumn{1}{c|}{} && CNN & $\textnormal{0.57}$ & $\textnormal{0.38}$ & $\textnormal{0.32}$ & \multicolumn{1}{c|}{$\textnormal{0.40}$} &
    $\fbox{\textnormal{0.19}}$ & $\textnormal{0.25}$ & $\textnormal{0.35}$ & $\textnormal{0.24}$ & $\textnormal{0.33}$ & \multicolumn{1}{c}{$\textnormal{0.49}$} \\
    \multicolumn{1}{c|}{} && WP & $\textnormal{0.45}$ & $\textnormal{0.35}$ & $\textnormal{0.40}$ & \multicolumn{1}{c|}{$\textnormal{0.38}$} &
    $\textnormal{0.42}$ & $\textnormal{0.36}$ & $\fbox{\textnormal{0.16}}$ & $\textnormal{0.40}$ & $\textnormal{0.41}$ & \multicolumn{1}{c}{$\textnormal{0.48}$} \\ \cline{2-13}

    \multicolumn{1}{c|}{} & \multirow{3}{*}{t$\rightarrow$a} & LSTM & $\textnormal{0.70}$ & $\textnormal{0.69}$ & $\textnormal{0.66}$ & \multicolumn{1}{c|}{$\textnormal{0.69}$} &
    $\textnormal{0.45}$ & $\fbox{\textnormal{0.34}}$ & $\textnormal{0.55}$ & $\textnormal{0.42}$ & $\textnormal{0.48}$ & \multicolumn{1}{c}{$\textnormal{0.65}$} \\
    \multicolumn{1}{c|}{} && CNN & $\textnormal{0.68}$ & $\textnormal{0.56}$ & $\textnormal{0.60}$ & \multicolumn{1}{c|}{$\textnormal{0.64}$} &
    $\fbox{\textnormal{0.36}}$ & $\textnormal{0.46}$ & $\textnormal{0.56}$ & $\textnormal{0.44}$ & $\textnormal{0.51}$ & \multicolumn{1}{c}{$\textnormal{0.66}$} \\
    \multicolumn{1}{c|}{} && WP & $\textnormal{0.73}$ & $\textnormal{0.58}$ & $\textnormal{0.61}$ & \multicolumn{1}{c|}{$\textnormal{0.70}$} &
    $\textnormal{0.61}$ & $\textnormal{0.59}$ & $\fbox{\textnormal{0.35}}$ & $\textnormal{0.57}$ & $\textnormal{0.59}$ & \multicolumn{1}{c}{$\textnormal{0.68}$} \\ \hline

    \multicolumn{1}{c|}{\multirow{6}{*}{AB}} & \multirow{3}{*}{A$\rightarrow$B} & LSTM & $\textnormal{0.88}$ & $\fbox{\textnormal{0.18}}$ & $\textnormal{0.18}$ & \multicolumn{1}{c|}{$\textnormal{0.74}$} &
    $\textnormal{0.54}$ & $\textnormal{0.36}$ & $\textnormal{0.63}$ & $\textnormal{0.56}$ & $\textnormal{0.62}$ & \multicolumn{1}{c}{$\textnormal{0.74}$} \\
    \multicolumn{1}{c|}{} && CNN & $\textnormal{0.92}$ & $\textnormal{0.22}$ & $\fbox{\textnormal{0.16}}$ & \multicolumn{1}{c|}{$\textnormal{0.80}$} &
    $\textnormal{0.46}$ & $\textnormal{0.50}$ & $\textnormal{0.58}$ & $\textnormal{0.56}$ & $\textnormal{0.64}$ & \multicolumn{1}{c}{$\textnormal{0.76}$} \\
    \multicolumn{1}{c|}{} && WP & $\textnormal{0.73}$ & $\textnormal{0.45}$ & $\textnormal{0.26}$ & \multicolumn{1}{c|}{$\textnormal{0.67}$} &
    $\textnormal{0.50}$ & $\textnormal{0.30}$ & $\fbox{\textnormal{0.12}}$ & $\textnormal{0.37}$ & $\textnormal{0.45}$ & \multicolumn{1}{c}{$\textnormal{0.60}$} \\ \cline{2-13}
    
    \multicolumn{1}{c|}{} & \multirow{3}{*}{B$\rightarrow$A} & LSTM & $\textnormal{0.94}$ & $\textnormal{0.16}$ & $\fbox{\textnormal{0.11}}$ & \multicolumn{1}{c|}{$\textnormal{0.70}$} &
    $\textnormal{0.76}$ & $\textnormal{0.52}$ & $\textnormal{0.78}$ & $\textnormal{0.75}$ & $\textnormal{0.78}$ & \multicolumn{1}{c}{$\textnormal{0.89}$} \\
    \multicolumn{1}{c|}{} && CNN & $\textnormal{0.93}$ & $\textnormal{0.10}$ & $\fbox{\textnormal{0.01}}$ & \multicolumn{1}{c|}{$\textnormal{0.61}$} &
    $\textnormal{0.62}$ & $\textnormal{0.72}$ & $\textnormal{0.84}$ & $\textnormal{0.80}$ & $\textnormal{0.78}$ & \multicolumn{1}{c}{$\textnormal{0.87}$} \\
    \multicolumn{1}{c|}{} && WP & $\textnormal{0.90}$ & $\fbox{\textnormal{0.09}}$ & $\textnormal{0.19}$ & \multicolumn{1}{c|}{$\textnormal{0.68}$} &
    $\textnormal{0.81}$ & $\textnormal{0.68}$ & $\textnormal{0.31}$ & $\textnormal{0.68}$ & $\textnormal{0.78}$ & \multicolumn{1}{c}{$\textnormal{0.85}$} \\ \hline

    \multicolumn{1}{c|}{\multirow{6}{*}{TO}} & \multirow{3}{*}{T$\rightarrow$O} & LSTM & $\textnormal{0.86}$ & $\textnormal{0.95}$ & $\textnormal{0.64}$ & \multicolumn{1}{c|}{$\fbox{\textnormal{0.34}}$} &
    $\textnormal{0.76}$ & $\textnormal{0.43}$ & $\textnormal{0.76}$ & $\textnormal{0.59}$ & $\textnormal{0.66}$ & \multicolumn{1}{c}{$\textnormal{0.81}$} \\
    \multicolumn{1}{c|}{} && CNN & $\textnormal{0.60}$ & $\textnormal{0.88}$ & $\textnormal{0.60}$ & \multicolumn{1}{c|}{$\textnormal{0.46}$} &
    $\fbox{\textnormal{0.36}}$ & $\textnormal{0.52}$ & $\textnormal{0.57}$ & $\textnormal{0.53}$ & $\textnormal{0.55}$ & \multicolumn{1}{c}{$\textnormal{0.58}$} \\
    \multicolumn{1}{c|}{} && WP & $\textnormal{0.75}$ & $\textnormal{0.61}$ & $\textnormal{0.85}$ & \multicolumn{1}{c|}{$\textnormal{0.46}$} &
    $\textnormal{0.63}$ & $\textnormal{0.58}$ & $\fbox{\textnormal{0.14}}$ & $\textnormal{0.58}$ & $\textnormal{0.60}$ & \multicolumn{1}{c}{$\textnormal{0.67}$} \\ \cline{2-13}
    
    \multicolumn{1}{c|}{} & \multirow{3}{*}{O$\rightarrow$T} & LSTM & $\textnormal{0.77}$ & $\fbox{\textnormal{0.00}}$ & $\textnormal{0.01}$ & \multicolumn{1}{c|}{$\textnormal{0.23}$} &
    $\textnormal{0.61}$ & $\textnormal{0.27}$ & $\textnormal{0.58}$ & $\textnormal{0.40}$ & $\textnormal{0.51}$ & \multicolumn{1}{c}{$\textnormal{0.65}$} \\
    \multicolumn{1}{c|}{} && CNN & $\textnormal{0.67}$ & $\fbox{\textnormal{0.00}}$ & $\textnormal{0.01}$ & \multicolumn{1}{c|}{$\textnormal{0.68}$} &
    $\textnormal{0.50}$ & $\textnormal{0.57}$ & $\textnormal{0.58}$ & $\textnormal{0.57}$ & $\textnormal{0.58}$ & \multicolumn{1}{c}{$\textnormal{0.63}$} \\
    \multicolumn{1}{c|}{} && WP & $\textnormal{0.73}$ & $\textnormal{0.31}$ & $\fbox{\textnormal{0.09}}$ & \multicolumn{1}{c|}{$\textnormal{0.36}$} &
    $\textnormal{0.62}$ & $\textnormal{0.57}$ & $\textnormal{0.16}$ & $\textnormal{0.56}$ & $\textnormal{0.61}$ & \multicolumn{1}{c}{$\textnormal{0.64}$} \\
    
    \end{tabular}
    \captionsetup{justification=centering}
    \caption{Author profiling accuracies in two-class sentence-based datasets: best (lowest) results framed.}
    \label{tab:results-imitation-baseline}
  \end{center}
\end{table*}

\newpage

\appendix
\section{Raw results}
\label{appendix:raw_results}

In this appendix we include the raw results from our sentence-based experiments, as well as example transformations by \sysname and all baseline techniques.

\begin{table*}[b!]
  \begin{center}
    \begin{tabular}{l|c|ccccc}
    Technique & Direction & YG & BA & AB & TO \\ \hline

    \multirow{2}{*}{\shortstack[l]{Original \\ (no transformations)}} & $0 \rightarrow 1$ & $\textnormal{0.83}$ & $\textnormal{0.54}$ & $\textnormal{0.88}$ & $\textnormal{0.86}$ \\
    & $1 \rightarrow 0$ & $\textnormal{0.82}$ & $\textnormal{0.70}$ & $\textnormal{0.94}$ & $\textnormal{0.77}$ \\ \hline

    \multirow{2}{*}{\shortstack[l]{Grammatical \\ transformations}} & $0 \rightarrow 1$ & $\textnormal{0.72}$ & $\textnormal{0.48}$ & $\textnormal{0.69}$ & $\textnormal{0.80}$ \\
    & $1 \rightarrow 0$ & $\textnormal{0.70}$ & $\textnormal{0.62}$ & $\textnormal{0.85}$ & $\textnormal{0.68}$ \\ \hline

    \multirow{2}{*}{Simple rules} & $0 \rightarrow 1$ & $\textnormal{0.81}$ & $\textnormal{0.51}$ & $\textnormal{0.85}$ & $\textnormal{0.82}$ \\
    & $1 \rightarrow 0$ & $\textnormal{0.82}$ & $\textnormal{0.66}$ & $\textnormal{0.91}$ & $\textnormal{0.72}$ \\ \hline

    \multirow{2}{*}{PPDB} & $0 \rightarrow 1$ & $\textnormal{0.65}$ & $\textnormal{0.37}$ & $\textnormal{0.73}$ & $\textnormal{0.64}$ \\
    & $1 \rightarrow 0$ & $\textnormal{0.62}$ & $\textnormal{0.51}$ & $\textnormal{0.84}$ & $\textnormal{0.50}$ \\ \hline

    \multirow{2}{*}{WordNet} & $0 \rightarrow 1$ & $\textnormal{0.65}$ & $\textnormal{0.36}$ & $\textnormal{0.77}$ & $\textnormal{0.65}$ \\
    & $1 \rightarrow 0$ & $\textnormal{0.70}$ & $\textnormal{0.48}$ & $\textnormal{0.80}$ & $\textnormal{0.52}$ \\ \hline

    \multirow{2}{*}{Typos} & $0 \rightarrow 1$ & $\textnormal{0.79}$ & $\textnormal{0.39}$ & $\textnormal{0.79}$ & $\textnormal{0.82}$ \\
    & $1 \rightarrow 0$ & $\textnormal{0.82}$ & $\textnormal{0.65}$ & $\textnormal{0.74}$ & $\textnormal{0.65}$ \\
    
    \end{tabular}
    \captionsetup{justification=centering}
    \caption{Author profiling accuracies with individual \sysname modules (LSTM profiler, \sysname-LSTM variant). \\
    Class $0$: $\{$female, adult, Alice, Trump$\}$; Class $1$: $\{$male, teen, Bob, Obama$\}$}
    \label{tab:results-parchoice-modules-raw}
  \end{center}
\end{table*}

\noindent{\textbf{Two-class experiments:}}
Table \ref{tab:results-imitation-baseline} displays two-class results for the original test sets and each style transfer direction on every style transfer technique, measured with all three author profilers (LSTM, CNN, WP).
Table \ref{tab:results-parchoice-modules-raw} shows corresponding results for each \sysname module applied separately, using the \sysname-LSTM variant and the LSTM profiler.

\begin{table*}[h!]
  \begin{center}
    \begin{tabular}{c|c|c|cc|cc|cc|cc|cc|}
    
    \multicolumn{1}{c|}{\multirow{3}{*}{\textbf{Source author}}} & \multirow{3}{*}{\textbf{Technique}} & \multirow{3}{*}{\shortstack[c]{\textbf{Author} \\ \textbf{profiler}}} & \multicolumn{10}{c}{\textbf{Target author; profiler accuracies with imitated test sets}} \\ \cline{4-13}
    \multicolumn{1}{c|}{} &&& \multicolumn{2}{c|}{A1} & \multicolumn{2}{c|}{A2} & \multicolumn{2}{c|}{A3} & \multicolumn{2}{c|}{A4} & \multicolumn{2}{c}{A5} \\
    \multicolumn{1}{c|}{} &&& s & t & s & t & s & t & s & t & s & \multicolumn{1}{c}{t} \\ \hline	
    
    \multicolumn{1}{c|}{\multirow{4}{*}{A1}} & \multirow{2}{*}{\sysname-LSTM} & LSTM & $\textnormal{(0.75)}$ & $-$ & $\textnormal{0.27}$ & $\textnormal{\textbf{0.51}}$ & $\textnormal{0.28}$ & $\textnormal{\textbf{0.35}}$ & $\textnormal{0.26}$ & $\textnormal{\textbf{0.32}}$ & $\textnormal{0.27}$ & \multicolumn{1}{c}{$\textnormal{\textbf{0.30}}$} \\
    
    \multicolumn{1}{c|}{} && CNN & $\textnormal{(0.82)}$ & $-$ & $\textnormal{0.51}$ & $\textnormal{0.28}$ & $\textnormal{0.52}$ & $\textnormal{0.25}$ & $\textnormal{0.45}$ & $\textnormal{0.26}$ & $\textnormal{0.52}$ & \multicolumn{1}{c}{$\textnormal{0.21}$} \\ \cline{2-13}

    \multicolumn{1}{c|}{} & \multirow{2}{*}{\sysname-CNN} & LSTM & $\textnormal{(0.75)}$ & $-$ & $\textnormal{0.43}$ & $\textnormal{0.32}$ & $\textnormal{0.43}$ & $\textnormal{0.22}$ & $\textnormal{0.42}$ & $\textnormal{0.14}$ & $\textnormal{0.43}$ & \multicolumn{1}{c}{$\textnormal{0.14}$} \\

    \multicolumn{1}{c|}{} && CNN & $\textnormal{(0.82)}$ & $-$ & $\textnormal{0.37}$ & $\textnormal{\textbf{0.39}}$ & $\textnormal{0.37}$ & $\textnormal{0.35}$ & $\textnormal{0.38}$ & $\textnormal{0.19}$ & $\textnormal{0.37}$ & \multicolumn{1}{c}{$\textnormal{0.33}$} \\ \hline

    \multicolumn{1}{c|}{\multirow{4}{*}{A2}} & \multirow{2}{*}{\sysname-LSTM} & LSTM & $\textnormal{0.42}$ & $\textnormal{0.40}$ & $\textnormal{(0.81)}$ & $-$ & $\textnormal{0.47}$ & $\textnormal{0.21}$ & $\textnormal{0.41}$ & $\textnormal{0.31}$ & $\textnormal{0.44}$ & \multicolumn{1}{c}{$\textnormal{0.15}$} \\

    \multicolumn{1}{c|}{} && CNN & $\textnormal{0.53}$ & $\textnormal{0.33}$ & $\textnormal{(0.82)}$ & $-$ & $\textnormal{0.59}$ & $\textnormal{0.13}$ & $\textnormal{0.48}$ & $\textnormal{0.28}$ & $\textnormal{0.59}$ & \multicolumn{1}{c}{$\textnormal{0.12}$} \\ \cline{2-13}

    \multicolumn{1}{c|}{} & \multirow{2}{*}{\sysname-CNN} & LSTM & $\textnormal{0.61}$ & $\textnormal{0.18}$ & $\textnormal{(0.81)}$ & $-$ & $\textnormal{0.62}$ & $\textnormal{0.12}$ & $\textnormal{0.61}$ & $\textnormal{0.13}$ & $\textnormal{0.62}$ & \multicolumn{1}{c}{$\textnormal{0.07}$} \\

    \multicolumn{1}{c|}{} && CNN & $\textnormal{0.46}$ & $\textnormal{0.29}$ & $\textnormal{(0.82)}$ & $-$ & $\textnormal{0.46}$ & $\textnormal{0.26}$ & $\textnormal{0.46}$ & $\textnormal{0.19}$ & $\textnormal{0.46}$ & \multicolumn{1}{c}{$\textnormal{0.25}$} \\ \hline

    \multicolumn{1}{c|}{\multirow{4}{*}{A3}} & \multirow{2}{*}{\sysname-LSTM} & LSTM & $\textnormal{0.25}$ & $\textnormal{\textbf{0.57}}$ & $\textnormal{0.24}$ & $\textnormal{\textbf{0.51}}$ & $\textnormal{(0.64)}$ & $-$ & $\textnormal{0.24}$ & $\textnormal{\textbf{0.37}}$ & $\textnormal{0.25}$ & \multicolumn{1}{c}{$\textnormal{\textbf{0.27}}$} \\

    \multicolumn{1}{c|}{} && CNN & $\textnormal{0.30}$ & $\textnormal{\textbf{0.48}}$ & $\textnormal{0.39}$ & $\textnormal{0.30}$ & $\textnormal{(0.76)}$ & $-$ & $\textnormal{0.36}$ & $\textnormal{0.32}$ & $\textnormal{0.38}$ & \multicolumn{1}{c}{$\textnormal{0.24}$} \\ \cline{2-13}

    \multicolumn{1}{c|}{} & \multirow{2}{*}{\sysname-CNN} & LSTM & $\textnormal{0.38}$ & $\textnormal{0.30}$ & $\textnormal{0.36}$ & $\textnormal{0.28}$ & $\textnormal{(0.64)}$ & $-$ & $\textnormal{0.43}$ & $\textnormal{0.15}$ & $\textnormal{0.43}$ & \multicolumn{1}{c}{$\textnormal{0.14}$} \\

    \multicolumn{1}{c|}{} && CNN & $\textnormal{0.28}$ & $\textnormal{\textbf{0.43}}$ & $\textnormal{0.29}$ & $\textnormal{\textbf{0.42}}$ & $\textnormal{(0.76)}$ & $-$ & $\textnormal{0.30}$ & $\textnormal{0.27}$ & $\textnormal{0.28}$ & \multicolumn{1}{c}{$\textnormal{\textbf{0.41}}$} \\ \hline
    
    \multicolumn{1}{c|}{\multirow{4}{*}{A4}} & \multirow{2}{*}{\sysname-LSTM} & LSTM & $\textnormal{0.30}$ & $\textnormal{\textbf{0.47}}$ & $\textnormal{0.28}$ & $\textnormal{\textbf{0.44}}$ & $\textnormal{0.30}$ & $\textnormal{0.29}$ & $\textnormal{(0.71)}$ & $-$ & $\textnormal{0.28}$ & \multicolumn{1}{c}{$\textnormal{\textbf{0.32}}$} \\

    \multicolumn{1}{c|}{} && CNN & $\textnormal{0.50}$ & $\textnormal{0.35}$ & $\textnormal{0.52}$ & $\textnormal{0.23}$ & $\textnormal{0.53}$ & $\textnormal{0.11}$ & $\textnormal{(0.79)}$ & $-$ & $\textnormal{0.46}$ & \multicolumn{1}{c}{$\textnormal{0.24}$} \\ \cline{2-13}

    \multicolumn{1}{c|}{} & \multirow{2}{*}{\sysname-CNN} & LSTM & $\textnormal{0.55}$ & $\textnormal{0.18}$ & $\textnormal{0.52}$ & $\textnormal{0.23}$ & $\textnormal{0.57}$ & $\textnormal{0.16}$ & $\textnormal{(0.71)}$ & $-$ & $\textnormal{0.55}$ & \multicolumn{1}{c}{$\textnormal{0.12}$} \\

    \multicolumn{1}{c|}{} && CNN & $\textnormal{0.31}$ & $\textnormal{0.29}$ & $\textnormal{0.31}$ & $\textnormal{\textbf{0.34}}$ & $\textnormal{0.31}$ & $\textnormal{\textbf{0.33}}$ & $\textnormal{(0.79)}$ & $-$ & $\textnormal{0.31}$ & \multicolumn{1}{c}{$\textnormal{\textbf{0.42}}$} \\ \hline

    \multicolumn{1}{c|}{\multirow{4}{*}{A5}} & \multirow{2}{*}{\sysname-LSTM} & LSTM & $\textnormal{0.36}$ & $\textnormal{\textbf{0.42}}$ & $\textnormal{0.35}$ & $\textnormal{\textbf{0.36}}$ & $\textnormal{0.35}$ & $\textnormal{\textbf{0.36}}$ & $\textnormal{0.33}$ & $\textnormal{\textbf{0.41}}$ & $\textnormal{(0.64)}$ & \multicolumn{1}{c}{$-$} \\

    \multicolumn{1}{c|}{} && CNN & $\textnormal{0.57}$ & $\textnormal{0.30}$ & $\textnormal{0.63}$ & $\textnormal{0.15}$ & $\textnormal{{0.71}}$ & $\textnormal{0.12}$ & $\textnormal{0.56}$ & $\textnormal{0.26}$ & $\textnormal{(0.75)}$ & \multicolumn{1}{c}{$-$} \\ \cline{2-13}

    \multicolumn{1}{c|}{} & \multirow{2}{*}{\sysname-CNN} & LSTM & $\textnormal{0.52}$ & $\textnormal{0.21}$ & $\textnormal{0.51}$ & $\textnormal{0.20}$ & $\textnormal{0.51}$ & $\textnormal{0.14}$ & $\textnormal{0.51}$ & $\textnormal{0.14}$ & $\textnormal{(0.64)}$ & \multicolumn{1}{c}{$-$} \\

    \multicolumn{1}{c|}{} && CNN & $\textnormal{0.45}$ & $\textnormal{0.34}$ & $\textnormal{0.48}$ & $\textnormal{0.27}$ & $\textnormal{0.45}$ & $\textnormal{0.27}$ & $\textnormal{0.46}$ & $\textnormal{0.21}$ & $\textnormal{(0.75)}$ & \multicolumn{1}{c}{$-$} \\
    
    \end{tabular}
    \captionsetup{justification=centering}
    \caption{Five-class author imitation results (s = source author accuracy, t = target author accuracy).\\
    Successful imitation in bold (target author accuracy $>$ source author accuracy).}
    \label{tab:results-multiclass-raw}
  \end{center}
\end{table*}

\begin{table*}[t!]
  \begin{center}
    \begin{tabular}{l|l|l}
    
    \multirow{5}{*}{\shortstack[l]{YG \\ (female$\rightarrow$male)}} & Original & \textnormal{the dinner portions are huge .} \\
    & CAE & \textnormal{the drinks are \$ $\textnormal{00}$ .} \\
    & BT & \textnormal{the rooms are great .} \\
    & A$^4$NT & \textnormal{the dinner portions are ultra .} \\
    & \sysname & \textnormal{the supper shares are tremendous .} \\ \hline

    \multirow{5}{*}{\shortstack[l]{BA \\ (adult$\rightarrow$teen)}} & Original & \textnormal{you feel like killing them but then again they are protected .} \\
    & CAE & \textnormal{you feel like then you are them .} \\
    & BT & \textnormal{eddy you want to see them , but now they are prot\'{e}g\'{e}s .} \\
    & A$^4$NT & \textnormal{you feel like killing them but then again they are .} \\
    & \sysname & \textnormal{you feel like popping them but then again theyre been safeguarded .} \\ \hline

    \multirow{5}{*}{\shortstack[l]{AB \\ (Alice$\rightarrow$Bob)}} & Original & \textnormal{we are so useless when it comes to bugs- its ridiculous !} \\
    & CAE & \textnormal{so yeah we went to <unk> it 's <unk> ... .} \\
    & BT & \textnormal{i was fattoria nous , , and de ... ..} \\
    & A$^4$NT & \textnormal{we are so far when it comes to me !} \\
    & \sysname & \textnormal{were so ineffectual when it is about microbes its farcical .} \\ \hline

    \multirow{5}{*}{\shortstack[l]{TO \\ (Obama$\rightarrow$Trump)}} & Original & \textnormal{i can tell you .} \\
    & CAE & \textnormal{we 're going .} \\
    & BT & \textnormal{i 's n't .} \\
    & A$^4$NT & \textnormal{i can you disagree} \\
    & \sysname & \textnormal{you might well be told by me .} \\ \hline \hline
    
    \multirow{5}{*}{\shortstack[l]{EBG$_5$}} & Original & \textnormal{They also tend to be somewhat adapted to fire of varying frequencies.} \\
    & PAN2016 & \textnormal{They also tended to be somewhat tailor to fire of varying frequencies.} \\
    & Mutant-X & \textnormal{They also tend to be somewhat adapted to fireball of aforementioned frequencies.} \\
    & \sysname & \textnormal{they also tend to be somewhat adjusted to a fire of differing frequencies .} \\
    & \sysname{}+Mutant-X & \textnormal{they also tend to continue somewhat adjusted to a fire of differing frequencies .} \\
        
    \end{tabular}
    \captionsetup{justification=centering}
    \caption{Style transfer examples (\sysname variant in YG/BA/AB/TO: \sysname-LR$_\textit{s}$).}
    \label{tab:results-examples}
  \end{center}  
\end{table*}

\noindent{\textbf{Five-class experiments:}}
Imitation results for each of the $20$ author pairs from the five-class experiments are collected in Table \ref{tab:results-multiclass-raw}.
Successful target author imitation results are written in bold (higher target author accuracy than source author accuracy).

\noindent{\textbf{Example transformations:}}
Table \ref{tab:results-examples} shows example transformations by \sysname, the encoder-decoder baselines (rows $1-20$) and the rule-based baselines (rows $21-25$).

\end{document}